\documentclass[10pt, a4paper, twocolumn, teaser, showabstract]{naverlabseurope}

\usepackage{tikz,pgfplots,tkz-kiviat}
\usepackage{arrayjobx}
\usepackage{graphicx}
\usepackage{caption}
\usepackage{subcaption}
\usepackage{booktabs}
\usepackage{multirow}
\usepackage{gensymb}
\usepackage{etoc}
\usepackage{pifont}
\usepackage{enumitem}
\usepackage{xspace}
\usepackage{adjustbox}
\usepackage{makecell}
\usepackage[normalem]{ulem}
\usepackage{wrapfig}
\usepackage{microtype}
\usepackage{appendix}

\usepackage{geometry}
\usepackage[capitalize]{cleveref}

\newcounter{rownumbers}
\newcounter{rownumbersft}
\renewcommand\rownum{{\color{darkgray} \stepcounter{rownumbers}\arabic{rownumbers}.}}
\newcommand\rownumft{{\color{darkgray} \stepcounter{rownumbersft}\arabic{rownumbersft}.}}

\crefname{section}{Sec.}{Secs.}
\Crefname{section}{Sec.}{Secs.}
\Crefname{table}{Tab.}{Tabs.}
\crefname{table}{Tab.}{Tabs.}
\Crefname{figure}{Fig.}{Figs.}
\crefname{figure}{Fig.}{Figs.}
\Crefname{equation}{Eq.}{Eqs.}
\crefname{equation}{Eq.}{Eqs.}
\crefname{appendix}{Appendix}{Appendices}

\makeatletter
\newcommand{\CapitalizeFirst}[1]{
  \expandafter\@CapitalizeFirst\expandafter{#1}
}
\def\@CapitalizeFirst#1{\MakeUppercase{\@car#1\@nil}\@cdr#1\@nil}
\makeatother

\makeatletter
\DeclareRobustCommand\onedot{\futurelet\@let@token\@onedot}
\def\@onedot{\ifx\@let@token.\else.\null\fi\xspace}
\def\eg{\emph{e.g}\onedot} 
\def\ie{\emph{i.e}\onedot} 
 
 \def\vs{\emph{vs}\onedot}
 
\def\etal{\emph{et al}\onedot}

\makeatother

\makeatletter
\renewcommand{\paragraph}[1]{
  \ifnum\pagetotal=\z@
    \noindent\textbf{#1}
  \else
    \vspace{2pt}\noindent\textbf{#1}
  \fi
}
\makeatother

\definecolor{tabdefault}{gray}{0.9}
\definecolor{lightgray}{gray}{0.93}
\newcommand{\insight}[1]{
\par\noindent\colorbox{lightgray}{
\begin{minipage}{0.97\linewidth}\vspace{1pt} $\blacktriangleright$ \emph{#1}\end{minipage}}
}

\newcommand{\supp}{supplementary material\xspace}

\newcommand{\dino}{DINO\xspace}
\newcommand{\ibot}{iBOT\xspace}
\newcommand{\deit}{DeiT-III\xspace}
\newcommand{\dbot}{dBOT-ft\xspace}

\newcommand{\dinodist}{DINO-distill\xspace}

\newcommand{\deitdist}{DeiT-III-distill\xspace}
\newcommand{\dbotdist}{dBOT-ft-distill\xspace}

\newcommand{\tdrop}{{\em tdrop}\xspace}
\newcommand{\dpshort}{LP\xspace}
\newcommand{\dplong}{ladder of projectors\xspace}

\newcommand{\divine}{UNIC\xspace}
\newcommand{\unic}{\divine}
\newcommand{\divineb}{\divine-${base}$\xspace}
\newcommand{\divinet}{\divine-${tdrop}$\xspace}
\newcommand{\divinedp}{\divine-${\dpshort}$\xspace}

\definecolor{tabsecond}{rgb}{0.8, 1, 0.8}
\definecolor{tabthird}{rgb}{0.88, 1, 0.88}
\definecolor{tabfirst}{rgb}{0.5, 1, 0.5}
\definecolor{tabtop}{rgb}{0.2, 1, 0.2}
\definecolor{tablast}{rgb}{1, 0.5, 0.5}
\definecolor{tablastest}{rgb}{1, 0.2, 0.2}
\definecolor{tabsecondlast}{rgb}{1, 0.8, 0.8}

\newcommand{\gradinggrad}{35}
\newcommand{\ok}[1]{{\cellcolor{tabthird!\gradinggrad}{#1}}}
\newcommand{\good}[1]{{\cellcolor{tabsecond!\gradinggrad}{#1}}}
\newcommand{\better}[1]{{\cellcolor{tabfirst!\gradinggrad}{#1}}}
\newcommand{\best}[1]{{\cellcolor{tabtop!\gradinggrad}{#1}}}
\newcommand{\degradesmost}[1]{{\cellcolor{tablastest!\gradinggrad}{#1}}}
\newcommand{\degrades}[1]{{\cellcolor{tablast!\gradinggrad}{#1}}}
\newcommand{\degradesless}[1]{{\cellcolor{tabsecondlast!\gradinggrad}{#1}}}

\newcommand{\gain}[1]{\textbf{\color{OliveGreen}{{$\uparrow$}{#1}}}}
\newcommand{\loss}[1]{\textbf{\color{BrickRed}{{$\downarrow$}{#1}}}}

\newcommand{\imsd}[1]{\textbf{\color{BrickRed}{{#1}}}}

\definecolor{teal}{RGB}{41,120,108}

\usepackage{amsmath,amsfonts,bm}

\def\eqref#1{equation~\ref{#1}}

\def\1{\bm{1}}

\def\vs{{\bm{s}}}
\def\vt{{\bm{t}}}

\def\vx{{\bm{x}}}
\def\vy{{\bm{y}}}
\def\vz{{\bm{z}}}

\DeclareMathAlphabet{\mathsfit}{\encodingdefault}{\sfdefault}{m}{sl}
\SetMathAlphabet{\mathsfit}{bold}{\encodingdefault}{\sfdefault}{bx}{n}

\def\gL{{\mathcal{L}}}

\def\gP{{\mathcal{P}}}

\def\gS{{\mathcal{S}}}
\def\gT{{\mathcal{T}}}

\usepackage{pgfplots, pgfplotstable}
\pgfplotsset{compat=newest}
\usepgfplotslibrary{fillbetween}
\usetikzlibrary{shapes.geometric}

\newcommand{\dinoc}{RedOrange}
\newcommand{\deitc}{PineGreen}
\newcommand{\dbotc}{MidnightBlue}

\newcommand{\divinec}{BrickRed}
\newcommand{\divinebc}{\divinec}
\newcommand{\divinetc}{\divinec}
\newcommand{\divinemlc}{\divinec}

\newcommand{\fourteachersc}{\divinec}

\newcommand{\dteacherc}{Orange}
\newcommand{\bteacherc}{CornflowerBlue}

\newcommand{\teacherareac}{Gray!15}

\newcommand{\utilheight}{5.0cm}
\newcommand{\plothightwidth}{5cm}

\newcommand{\mrksz}{3}

\pgfdeclareplotmark{divinebm}{
    \node[star, star points=4, star point ratio=2.05, draw=black, solid, fill=\divinebc, minimum width=2.2*\mrksz, inner sep=0pt, outer sep=0pt, anchor=center] {};
}

\pgfdeclareplotmark{ddivinebm}{
    \node[star, star points=6, star point ratio=2.05, draw=black, solid, fill=\dteacherc, minimum width=2.2*\mrksz, inner sep=0pt, outer sep=0pt, anchor=center] {};
}
\pgfdeclareplotmark{bdivinebm}{
    \node[star, star points=6, star point ratio=2.05, draw=black, solid, fill=\bteacherc, minimum width=2.2*\mrksz, inner sep=0pt, outer sep=0pt, anchor=center] {};
}
\pgfdeclareplotmark{ddivinetm}{
    \node[star, star points=6, star point ratio=2.05, draw=black, solid, fill=\dteacherc, minimum width=2.2*\mrksz, inner sep=0pt, outer sep=0pt, anchor=center] {};
}
\pgfdeclareplotmark{bdivinetm}{
    \node[star, star points=6, star point ratio=2.05, draw=black, solid, fill=\bteacherc, minimum width=2.2*\mrksz, inner sep=0pt, outer sep=0pt, anchor=center] {};
}
\pgfdeclareplotmark{ddivinedm}{
    \node[star, star points=6, star point ratio=2.05, draw=black, solid, fill=\dteacherc, minimum width=2.2*\mrksz, inner sep=0pt, outer sep=0pt, anchor=center] {};
}
\pgfdeclareplotmark{bdivinedm}{
    \node[star, star points=6, star point ratio=2.05, draw=black, solid, fill=\bteacherc, minimum width=2.2*\mrksz, inner sep=0pt, outer sep=0pt, anchor=center] {};
}
\pgfdeclareplotmark{ddivinem}{
    \node[star, star points=5, star point ratio=2.05, draw=black, solid, fill=\dteacherc, minimum width=2.2*\mrksz, inner sep=0pt, outer sep=0pt, anchor=center] {};
}
\pgfdeclareplotmark{bdivinem}{
    \node[star, star points=6, star point ratio=2.05, draw=black, solid, fill=\bteacherc, minimum width=2.2*\mrksz, inner sep=0pt, outer sep=0pt, anchor=center] {};
}

\pgfdeclareplotmark{fdivinem}{
    \node[star, star points=4, star point ratio=2.05, draw=black, solid, fill=\fourteachersc, minimum width=2.2*\mrksz, inner sep=0pt, outer sep=0pt, anchor=center] {};
}

\pgfdeclareplotmark{divinetm}{
    \node[star, star points=5, star point ratio=2.25, draw=black, solid, fill=\divinetc, minimum width=2.2*\mrksz, inner sep=0pt, outer sep=0pt, anchor=center] {};
}

\pgfdeclareplotmark{divinemlm}{
    \node[star, star points=6, star point ratio=2.25, draw=black, solid, fill=\divinemlc, minimum width=2.2*\mrksz, inner sep=0pt, outer sep=0pt, anchor=center] {};
}

\pgfdeclareplotmark{divinem}{
    \node[star, star points=8, star point ratio=2.25, draw=black, solid, fill=\divinec, minimum width=2.2*\mrksz, inner sep=0pt, outer sep=0pt, anchor=center] {};
}

\pgfplotsset{
    dino/.style={only marks, fill=\dteacherc, mark=pentagon*, mark size=\mrksz},
    deit/.style={only marks, fill=\dteacherc, mark=square*, mark size=\mrksz},
    dbot/.style={only marks, fill=\bteacherc, mark=diamond*, mark size=\mrksz},
    ibot/.style={only marks, fill=\bteacherc, mark=triangle*, mark size=\mrksz},
    ddivineb/.style={only marks, fill=\dteacherc, mark=ddivinebm, mark size=\mrksz},
    bdivineb/.style={only marks, fill=\bteacherc, mark=bdivinebm, mark size=\mrksz},
    ddivinet/.style={only marks, fill=\dteacherc, mark=ddivinetm, mark size=\mrksz},
    bdivinet/.style={only marks, fill=\bteacherc, mark=bdivinetm, mark size=\mrksz},
    ddivined/.style={only marks, fill=\dteacherc, mark=ddivinedm, mark size=\mrksz},
    bdivined/.style={only marks, fill=\bteacherc, mark=bdivinedm, mark size=\mrksz},
    ddivine/.style={only marks, fill=\dteacherc, mark=ddivinem, mark size=\mrksz},
    bdivine/.style={only marks, fill=\bteacherc, mark=bdivinem, mark size=\mrksz},
    dinost/.style={solid, only marks, draw=black, fill=\dinoc, mark=triangle*, mark size=\mrksz},
    deitst/.style={only marks, draw=black, fill=\deitc, mark=triangle*, mark size=\mrksz},
    dbotst/.style={only marks, fill=\dbotc, mark=triangle*, mark size=\mrksz},
    divineb/.style={only marks, draw=black, fill=\divinebc, mark=divinebm, mark size=\mrksz},
    divinet/.style={only marks, draw=black, fill=\divinetc, mark=divinetm, mark size=\mrksz},
    divineml/.style={only marks, draw=black, fill=\divinemlc, mark=divinemlm, mark size=\mrksz},
    divine/.style={only marks, draw=black, fill=\divinec, mark=divinem, mark size=\mrksz},
    fdivine/.style={only marks, draw=black, fill=\fourteachersc, mark=fdivinem, mark size=\mrksz},
    teachutil/.style={gray, mark=*, mark size=\mrksz-2},
    fdivineline/.style={thick, \fourteachersc, mark=*, mark size=\mrksz-2},
}

\newcommand{\leg}[1]{\addlegendentry{#1}}

\tikzset{every mark/.append style={solid}}
\pgfplotsset{
    grid=none,
    width=\linewidth, try min ticks=5,
    legend cell align=left,
    legend style={fill opacity=0.8},
	ylabel near ticks,
    xlabel near ticks,
    every tick label/.append style={font=\footnotesize},
}

\makeatletter
\usepackage{trimspaces}
\def\trimspace#1{\trim@spaces@in{#1}}
\newcommand\HUGE{\@setfontsize\Huge{31}{39}}
\makeatother

\etocdepthtag.toc{mtchapter}
\etocsettagdepth{mtappendix}{none}

\correspondingauthor{[bulent.sariyildiz,yannis.kalantidis]@naverlabs.com}
\contributions{}
\title{UNIC: Universal Classification Models \\ via Multi-teacher Distillation}
\titlerunning{UNIC: Universal Classification Models via Multi-teacher Distillation}

\affiliations{NAVER LABS Europe}

\authors{Mert B\"{u}lent Sar{\i}y{\i}ld{\i}z \authsep Philippe Weinzaepfel \authsep Thomas Lucas \authsep Diane Larlus \authsep Yannis Kalantidis}
\website{https://europe.naverlabs.com/unic}
\websiteref{\href{https://europe.naverlabs.com/unic}}

\begin{abstract}

Pretrained models have become a commodity and offer strong results on a broad range of tasks.
In this work, we focus on classification and seek to learn a unique encoder able to take from several
complementary
pretrained models.
We aim at even \textit{stronger generalization} across a variety of classification tasks.
We propose to learn such an encoder via multi-teacher distillation.
We first thoroughly analyze standard distillation when driven by multiple strong teachers with complementary strengths. Guided by this analysis, we gradually propose improvements to the basic distillation setup.
Among those, we enrich the architecture of the encoder with a ladder of expendable projectors, which increases the impact of intermediate features during distillation, and we introduce teacher dropping, a regularization mechanism that better balances the teachers' influence.
Our final distillation strategy leads to student models of the same capacity as any of the teachers,
while retaining or improving upon the performance of the best teacher for each task.

\end{abstract}

\begin{document}

\maketitle

\section{Introduction}
\label{sec:intro}

\looseness=-1
Recent years have
witnessed the rise
of many pretrained models~\cite{caron2021emerging,touvron2021deit,zhou2022ibot}. They often share the same architecture
and sometimes even the same training data.
They generalize to a broad range of tasks, but may
particularly excel at specific visual recognition scenarios
depending
on the selected learning strategy.
Self-supervised learning models~\cite{chen2020simclr,chen2021simsiam,caron2021emerging} shine in transfer learning, \ie generalization to novel classes, while models trained with masked modeling techniques~\cite{he2022masked,zhou2022ibot}
are often better suited to
patch-level tasks.
Meanwhile, supervised learning~\cite{krizhevsky2012alexnet,dosovitskiy2021an} is still best
for specific classification tasks when labeled data is available during pretraining.

\looseness=-1
In this paper, our goal is to learn \textit{a universal encoder} capable of strong generalization across a broad spectrum of classification tasks.
More specifically, besides ImageNet classification~\cite{russakovsky2015ilsvrc}
-- the dataset on which all our teachers are trained and our students are distilled --
we are further interested in
the classification of novel classes, on new domains,
as well as dense prediction tasks such as semantic segmentation or depth estimation.
Our goal is
to learn \textit{a {single} encoder} that can be directly applied to all these tasks, out-of-the-box, without the need for any task-specific parameters besides a linear classifier
per classification task.

\begin{figure}[t]
    \begin{center}
    \vspace{10pt}
    \adjustbox{width=1.00\linewidth}{
        \centering
        \newarray\kivaxisitemlabels
\readarray{kivaxisitemlabels}{
&  &  &  &  & Depth estimation      &
&  &  &  &  & ImageNet-1K           &
&  &  &  &  & IN-V2                  &
&  &  &  &  & {\begin{tabular}{c}Domain \\ shift \end{tabular}}             &
&  &  &  &  & {\begin{tabular}{c}Concept \\ generalization \end{tabular}}                &
&  &  &  &  & Long-tail             &
&  &  &  &  & Fine-grained          &
&  &  &  &  & Segmentation         &
}

\newarray\kivaxisextralabelsxshift
\readarray{kivaxisextralabelsxshift}{
&  &   &   &  &  &
&  &   &   &  &  &
&  &   &   &  &  &
&  &   &   &  &  &
&  &   &   &  &  &
&  &   &   &  &  &
&  &   &   &  &  &
&  &   &   &  &   &
}
\newarray\kivaxisextralabels
\readarray{kivaxisextralabels}{
&  &   &   &  & \text{+}2.4\% &
&  &   &   &  & -0.2\% &
&  &   &   &  & \text{+}0.4\% &
&  &   &   &  & -0.6\%&
&  &   &   &  & \text{+}3.5\%&
&  &   &   &  & \text{+}2.6\%&
&  &   &   &  & \text{+}9.2\% &
&  &   &   &  & \text{+}8.2\%&
}

\dataheight=6

\newcommand{\kivcurrentlabel}[2]{\checkkivaxisitemlabels(#1,#2)\trimspace\cachedata \cachedata}
\newcommand{\kivcurrentextralabel}[2]{\checkkivaxisextralabels(#1,#2)\trimspace\cachedata \cachedata}
\newcommand{\kivcurrentextralabelxshift}[2]{\checkkivaxisextralabelsxshift(#1,#2)\trimspace\cachedata \cachedata}

\newcommand{\kivaxisnumbers}{8}
\newcommand{\kivcategorycounts}{{6,6,6,6,6,6,6,6}}
\newcommand{\kivlcmcatcount}{6}

\newcommand{\kivlattice}{
\pgfmathsetmacro{\kivaxisangle}{360/\kivaxisnumbers}
    \foreach \x in {1,...,\kivaxisnumbers}
    {   \foreach \y in {1,...,\kivlcmcatcount}
        {   \pgfmathsetmacro{\kivaxisstep}{6/\kivlcmcatcount}
            \draw[help lines,gray,line width=0.1mm] (\kivaxisangle*\x:\y*\kivaxisstep) -- (\kivaxisangle*\x+\kivaxisangle:\y*\kivaxisstep);
        }
    }
    \foreach \x in {1,...,\kivaxisnumbers}
    {   \draw[-,gray,line width=0.1mm] (0,0) -- (\kivaxisangle*\x:6cm);
        \pgfmathsetmacro{\kivaxissteps}{\kivcategorycounts[\x-1]}
        \pgfmathsetmacro{\kivaxisstep}{6/\kivcategorycounts[\x-1]}
        \foreach \y in {1,...,\kivaxissteps}
        {
            \pgfmathtruncatemacro{\kivlabelnumber}{\y}
            \pgfmathtruncatemacro{\ylabelpos}{\y+1}

        }

    }
}

\newcommand{\kivboxes}{
\pgfmathsetmacro{\kivaxisangle}{360/\kivaxisnumbers}
    \foreach \x in {1,...,\kivaxisnumbers}
    {   \foreach \y in {1,...,\kivlcmcatcount}
        {   \pgfmathsetmacro{\kivaxisstep}{6/\kivlcmcatcount}
        }
    }
    \foreach \x in {1,...,\kivaxisnumbers}
    {   \draw[-,gray,line width=0.1mm] (0,0) -- (\kivaxisangle*\x:6cm);
        \pgfmathsetmacro{\kivaxissteps}{\kivcategorycounts[\x-1]}
        \pgfmathsetmacro{\kivaxisstep}{6/\kivcategorycounts[\x-1]}
        \pgfmathtruncatemacro{\kivlabelnumber}{\kivaxissteps}
        \pgfmathtruncatemacro{\kivnewangle}{\kivaxisangle}
        \pgfmathtruncatemacro{\ypos}{\kivaxissteps+0.1}

        \node at (\kivnewangle*\x:\kivaxisstep*\ypos) [rectangle,draw=\divinec,fill=white] {{\huge \kivcurrentextralabel{\x}{\kivlabelnumber}}};

    }
}

\newcommand{\kivdatapoints}{}

\newcommand{\kivdata}[2]{
    \renewcommand{\kivdatapoints}{{#1}}
    \pgfmathsetmacro{\kivaxisangle}{360/\kivaxisnumbers}
    \pgfmathsetmacro{\kivcoordinate}{\kivdatapoints[0]*6}
    \draw[#2,line width=1.5mm] (\kivaxisangle:\kivcoordinate)
        \foreach \x in {1,...,\kivaxisnumbers}
        {
            -- (\kivaxisangle*\x:\kivdatapoints[\x-1]*6)
        }
    -- cycle;
}
\newcommand{\kivdatathin}[2]{
    \renewcommand{\kivdatapoints}{{#1}}
    \pgfmathsetmacro{\kivaxisangle}{360/\kivaxisnumbers}
    \pgfmathsetmacro{\kivcoordinate}{\kivdatapoints[0]*6}
    \draw[#2,line width=1mm] (\kivaxisangle:\kivcoordinate)
        \foreach \x in {1,...,\kivaxisnumbers}
        {
            -- (\kivaxisangle*\x:\kivdatapoints[\x-1]*6)
        }
    -- cycle;
}

\newcommand{\kivdashed}[2]{
    \renewcommand{\kivdatapoints}{{#1}}
    \pgfmathsetmacro{\kivaxisangle}{360/\kivaxisnumbers}
    \pgfmathsetmacro{\kivcoordinate}{\kivdatapoints[0]*6}
    \draw[#2,line width=1.5mm,dashed] (\kivaxisangle:\kivcoordinate)
        \foreach \x in {1,...,\kivaxisnumbers}
        {
            -- (\kivaxisangle*\x:\kivdatapoints[\x-1]*6)
        }
    -- cycle;
}

\newcommand{\kivshade}[2]{
    \renewcommand{\kivdatapoints}{{#1}}
    \pgfmathsetmacro{\kivaxisangle}{360/\kivaxisnumbers}
    \pgfmathsetmacro{\kivcoordinate}{\kivdatapoints[0]*6}
    \fill[opacity=0.1,#2] (\kivaxisangle:\kivcoordinate)
        \foreach \x in {1,...,\kivaxisnumbers}
        {
            -- (\kivaxisangle*\x:\kivdatapoints[\x-1]*6)
        }
    -- cycle;
        \foreach \x in {1,...,\kivaxisnumbers}
        {
            -- (\kivaxisangle*\x:\kivdatapoints[\x-1]*6)
        }
    -- cycle;
}

\begin{tikzpicture}
\kivlattice

\kivshade{0.6750, 0.6981, 0.6981, 0.3279, 0.8169, 0.8333, 0.8333, 0.5279}{black}
\kivshade{0.6062, 0.8247, 0.8243, 0.8333, 0.7813, 0.7383, 0.5544, 0.6215}{black}
\kivshade{0.8333, 0.7303, 0.7274, 0.3557, 0.8333, 0.8289, 0.8129, 0.8333}{black}
\kivshade{0.5168, 0.8333, 0.8333, 0.8014, 0.8305, 0.7714, 0.7551, 0.6461}{black}

\kivdatathin{0.6750, 0.6981, 0.6981, 0.3279, 0.8169, 0.8333, 0.8333, 0.5279}{black!100}
\kivdatathin{0.6062, 0.8247, 0.8243, 0.8333, 0.7813, 0.7383, 0.5544, 0.6215}{black!60}
\kivdatathin{0.8333, 0.7303, 0.7274, 0.3557, 0.8333, 0.8289, 0.8129, 0.8333}{black!80}
\kivdatathin{0.5168, 0.8333, 0.8333, 0.8014, 0.8305, 0.7714, 0.7551, 0.6461}{black!40}
\kivdata{0.8780, 0.8290, 0.8400, 0.8213, 0.8962, 0.8797, 1.0, 0.9810}{\divinec}

\kivboxes{}
\newcommand{\boxxspace}{4.7}
\newcommand{\boxyoffset}{-0.4}

\node[draw, black!70, minimum width=1cm, minimum height=0.07cm, fill,
label={[yshift=-14pt, xshift=72]{\Huge Teachers}}] at (\boxxspace,6.5-\boxyoffset) {};
\node[draw, \divinec, minimum width=1cm, minimum height=0.07cm, fill, label={[yshift=-14pt, xshift=60]{\Huge \unic}}] at (\boxxspace,5.5-\boxyoffset) {};
\draw (\boxxspace-0.7,7-\boxyoffset) rectangle (\boxxspace+4.5,5.1-\boxyoffset);

\node[label={[xshift=-50pt, yshift=20pt]{\Huge {\begin{tabular}{c}ImageNet-1K\end{tabular}}}}] at (1,5.5) {};
\node[label={[xshift=-1, yshift=-15pt]{\Huge {\begin{tabular}{c}Depth \\ estimation\end{tabular}}}}] at (7,2.5) {};
\node[label={[xshift=0, yshift=-15pt]{\Huge {\begin{tabular}{c}ImageNet-v2\end{tabular}}}}] at (-5,5) {};
\node[label={[xshift=30, yshift=-25pt]{\Huge {\begin{tabular}{c}Semantic \\ Segmentation\end{tabular}}}}] at (7,-2) {};
\node[label={[xshift=10, yshift=-23pt]{\Huge {\begin{tabular}{c}Domain \\Shift \end{tabular}}}}] at (-8,1) {};
\node[label={[xshift=50, yshift=-35pt]{\Huge {\begin{tabular}{c}Concept \\ Generalization\end{tabular}}}}] at (-8,-6) {};
\node[label={[xshift=-50pt, yshift=20pt]{\Huge {\begin{tabular}{c}ImageNet-1K\end{tabular}}}}] at (1,5.5) {};
\node[label={[xshift=-55pt, yshift=10pt]{\Huge {\begin{tabular}{c}Long-tail\end{tabular}}}}] at (5,-8) {};
\node[label={[xshift=12pt, yshift=30pt]{\Huge {\begin{tabular}{c}Fine-grained\end{tabular}}}}] at (7.5,-6.5) {};

\end{tikzpicture}

    }
    \end{center}
    \caption{{
        \textbf{Relative gains using our \unic} encoder distilled from four teachers (\dino, \deit, \ibot, \dbot), over the respective best teacher for each task. UNIC solves all classification tasks using a \textit{single encoder} and no task-specific parameters.}
    }
    \label{fig:spider}
\end{figure}
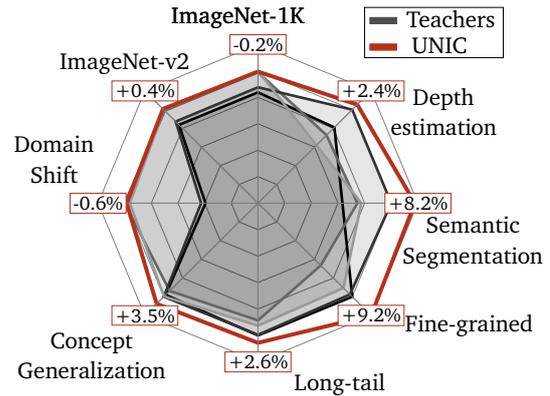

\looseness=-1
Our approach uses multi-teacher distillation, drawing on the strengths of various specialized teachers to train an encoder that seeks to match or surpass the best teacher
at each task.
We conduct a comprehensive analysis of the distillation process from multiple teachers,
evaluating our models on various tasks, including image-level classification on ImageNet-1K and 15 more transfer datasets, as well as patch-level classification tasks such as semantic segmentation and depth estimation.
We leverage our findings to gradually devise a method that shows improved generalization across multiple tasks and axes.
We
modify the input
of expandable projectors~\cite{chen2020simclr, chen2021simsiam, sariyildiz2023trex} (building what we call a \textit{\dplong}) so that
they
also act as information highways that propagate
signal from intermediate layers to the distillation loss in a more direct manner. We analyze learning dynamics across teachers and
further
propose \textit{teacher dropping}, an effective strategy for balancing the teachers' influence
in multi-teacher
distillation, resulting in significant
gains for the tasks
at which our distilled models were
otherwise underperforming.

\looseness=-1
With all of our improvements added to the basic multi-teacher distillation setup, we are able to train models that
exhibit strong generalization across a wide range of classification tasks
at the image and patch levels, either retaining or improving the performance of the best teacher.
As an example, we show in \Cref{fig:spider} that by distilling from four strong {ViT-Base models trained on ImageNet}
(\ie DINO~\cite{caron2021emerging}, DeiT-III~\cite{touvron2022deitiii}, iBOT~\cite{zhou2022ibot}, and dBOT-ft~\cite{liu2024dbot}) we are able to train a
\textit{universal encoder}
excelling at all considered tasks.
In our experimental study, we show that our findings further extend to the case of larger teachers like DINOv2~\cite{oquab2024dinov2} and MetaCLIP~\cite{xu2023metaclip} trained on arbitrary datasets.
{Finally,}
we
{study}
the way the distilled encoders utilize their weights:
first, by quantifying  performance drops after weights pruning, and second after reducing the dimension of the output feature space using PCA. These experiments show that distilled models have
lower redundancy in both their weights and their features.

\paragraph{Contributions.}
To summarize, we conduct a thorough analysis of multi-teacher distillation for ViT encoders and use our findings to improve the distillation process and generalization power of the student.
Among other simple but crucial modifications, we introduce improvements like \dplong and teacher dropping regularization that enable us to learn models which retain or improve the performance of the best teachers across many diverse tasks.
We refer to such models as \textbf{Uni}versal \textbf{C}lassification models or \textbf{\unic}.
We finally perform extensive evaluations along multiple axes of generalization and study the ways the resulting models make use of their weights and feature space.

\section{Related Work}\label{sec:relwork}

\looseness=-1
\paragraph{Knowledge distillation} (KD) {
was initially introduced as a model compression technique} \cite{bucilua2006model},
where the goal is to train a smaller student model from the output of a
teacher model~\cite{hinton2014distilling}.
While early work
focused on predicting the final outputs of a classification model,
the idea was rapidly extended to other forms of distillation, such as distilling intermediate
representations~\cite{ahn2019variational,heo2019knowledgetransfer,heo2019comprehensive,romero2015fitnets,zagoruyko2017paying,you2017learning,yim2017gift}.
These methods perform well but require
careful layer selection and loss balancing~\cite{heo2019comprehensive}.
In our work, instead of matching layer-wise representations between the student and teacher architectures,
we add shortcut connections
from intermediate layers of the student
to the loss of each teacher.

\paragraph{Multi-teacher knowledge distillation.}
KD can naturally be extended to an ensemble of teachers so that student can benefit from their potential complementarity.
While the final outputs of teachers trained for the same task can simply be averaged~\cite{asif2019ensemble,fukuda2017efficientkd,hinton2014distilling,you2017learning}, multi-teacher distillation with teachers trained for different tasks is more challenging.
UDON~\cite{ypsilantis2024udon} first trains domain-specialist teachers which are subsequently distilled in a student model using adaptive data sampling for balancing the different domains.
In \cite{tian2020contrastiverd},
contrastive learning is used for ensemble distillation
while \cite{shi2023hybrid} proposes a framework tailored for teachers trained with masked image modeling and contrastive learning.
But such approaches are not straightforward to extend to teachers learned differently.
Similarly, \cite{yao2023moma} combines self-supervised teachers from arbitrary heterogeneous pretext tasks.
\cite{ghiasi2021multi, clark2019bam,roth2024fantastic} focus on jointly utilizing  pseudo- and true labels for multi-teacher distillation. Roth \etal \cite{roth2024fantastic} formulate multi-teacher distillation as continual learning and further propose a novel method for data partitioning based on confidence.
Here we develop a more generic method for combining teachers, that is not limited to certain types of teachers
or losses, and, unlike~\cite{landgraf2024efficient, roth2024fantastic}, does not require labeled data, nor classifiers associated with each teacher for obtaining pseudo-labels.

\looseness=-1
\paragraph{Loss balancing}
is shown to be crucial in multi-task learning~\cite{chen2018gradnorm,kendall2018multi,hu2019learning,yun2023achievement}. Similar strategies to automatically balance losses have also been proposed for multi-teacher distillation~\cite{liu2020adaptive,fukuda2017efficientkd}. In \cite{hu2019learning},
adaptive loss weights inversely proportional to the average of each loss are introduced,
while~\cite{liu2020adaptive} learns instance-level teacher importance weights using ground-truth labels.
In \cite{fukuda2017efficientkd},
the random selection of one teacher per mini-batch is shown to help.
Our experiments show that
our
proposed generalized teacher dropping strategy leads to better models compared to~\cite{fukuda2017efficientkd,hu2019learning}.

\looseness=-1
\paragraph{Distilling from a ``foundation model''} like CLIP \cite{radford2021clip} or DINOv2 \cite{oquab2024dinov2}
is an effective approach for tasks with limited training data~\cite{marrie2024good,wei2022mvp,peng2023a}.
Distilling from \textit{multiple}
foundation models allows for more versatile students.
Recent works like AM-RADIO~\cite{ranzinger2023radio}, SAM-CLIP~\cite{wang2023sam}, and Open Vocabulary SAM~\cite{yuan2024open} combine the semantics captured by CLIP with the localization capabilities of models
like DINOv2~\cite{oquab2024dinov2} or SAM~\cite{kirillov2023segment}.
AM-RADIO~\cite{ranzinger2023radio} builds on the same
base setup as our study, but employs no loss balancing.
Another difference comes from the fact that
their student encoder is only a part of the final model:
AM-RADIO requires the teacher-specific projectors learned during distillation to also be used
at test time, effectively increasing the parameters of the encoder with task-specific ones.
Instead, our method performs well on multiple classification tasks \textit{out-of-the-box}, without any
additional parameters.

\looseness=-1
\paragraph{Combining models beyond distillation.}
Other ways to combine pretrained models have been proposed.
Works like \cite{wortsman2022model,rame2022diverse,matena2022merging,rame2023model,stoica2024zipit} explore different weight averaging strategies.
They typically only combine models that differ by their hyper-parameter configuration.
Aiming at generalization, \cite{ye2023merging} merges multiple ViTs, each specialized to a
classification task, into a single encoder that solves all classification tasks jointly, via a gating network.
Instead, our students are distilled from scratch, have a simple ViT architecture,
and tackle diverse tasks with simple linear probing.

\paragraph{Expendable projectors} are extra modules that act as buffers between the final encoder output and the
space where the loss is computed.
They have been successfully used for
both self-supervised~\cite{chen2020simclr, chen2021simsiam} and supervised learning~\cite{wang2022revisiting, sariyildiz2023trex}.
We extend this idea and add projectors during training to intermediate layers as well.
Roth \etal~\cite{roth2021simultaneous} use several such projectors of varied dimensionality for metric learning, but do not use features from intermediate layers.
Moreover, we use a specific set of projectors per teacher,
similar to
\cite{ranzinger2023radio, asif2019ensemble}.
This way, projectors become
\textit{loss-specific}, \ie they contribute to the loss for only one of the teachers.

\section{Improving multi-teacher distillation}
\label{sec:analysis}

\looseness=-1
In this section we first present the multi-teacher distillation setup we use as a basis for our analysis (\Cref{sec:basic_setup}) and a summary of our evaluation protocol
(\Cref{sec:protocol_summary}). We then delve into challenges
around multi-teacher distillation of ViT encoders (\Cref{sec:tokens}), and offer improvements to the basic setup to overcome them, like enhanced expendable teacher-specific projectors heads (\Cref{sec:projectors}) and strategies to more equally learn from all teachers
(\Cref{sec:balancing}).

\subsection{A basic distillation setup}
\label{sec:basic_setup}

\begin{figure*}[t!]
    \centering
    \includegraphics[width=\linewidth]{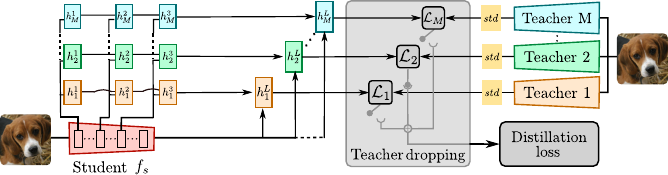} \\[-0.2cm]
    \caption{\textbf{Overview of our multi-teacher distillation setup.}
    {The same input image is fed to each teacher and to student.
    We employ feature standardization at the output of all teachers (\Cref{sec:tokens}), \textit{a ladder of expandable projectors} attached to student (\Cref{sec:projectors}) and \textit{teacher dropping regularization} to balance teachers (\Cref{sec:balancing}).
    The latter enables us to adaptively select a subset of teachers to contribute to the loss simply using loss magnitudes. We use dedicated projectors for the CLS and patch tokens (\Cref{sec:tokens}).}
    }
    \label{fig:overview}
\end{figure*}

\looseness=-1
Our goal is to distil $M$ \textit{teacher} models $\gT = \{\gT_1, \ldots, \gT_M\}$ into a student model $\gS$.
An overview is shown in Figure~\ref{fig:overview}.
Each teacher $t \in \gT$
{is a ViT\cite{dosovitskiy2021an} encoder that}
maps an image $\vx$ to a set of $d$-dimensional feature vectors $\vy_{t,i} = f_t(\vx ; i)$ for token $i$, {which} can either be one of the $H \times W$ patch tokens from $\gP$ or the global CLS token $c$.
We aim at learning the parameters $f_s$ of the student $\gS$, such that the output representations $\vz_i = f_s(\vx;i)$ excel at all the tasks that any of the teachers also shines at.

We append a \textit{projector head} $h_t$ per teacher to the student encoder's output which transforms each token into a teacher-specific representation $h_t(\vz_i)$. The loss for each teacher is then computed on
$h_t(\vz_i)$, the output of the corresponding projector head.
We
consider these projector heads as \textit{expendable}, \ie they are removed after distillation and are not part of the student encoder. Their goal is to assist the learning process, taking inspiration from
similar expendable projectors used in self-supervised~\cite{chen2020simclr} and supervised~\cite{wang2022revisiting,sariyildiz2023trex} representation learning.
We set projector heads to be Multi-Layer Perceptrons (MLPs) with two linear layers, GeLU non-linearity and hidden dimension of $d_h=4d$, where $d$ is the feature dimension; we analyze projectors further in the next sections.

We use the combination of two common distillation losses: cosine and smooth-$\ell_1$ (see \supp for details); the loss for token $i$ from teacher $t$ is given by:

\begin{equation}
    \gL_t(\vx;i) = \frac{ \gL^{cos}(h_t(\vz_i), \vy_{t,i}) + \gL^{s\ell_1}(h_t(\vz_i), \vy_{t,i}) }{2}.
    \label{eq:loss_cos_sl1}
\end{equation}

This loss is computed separately for the CLS and each of the patch tokens $\gP$.
To get the final loss, we sum losses from all teachers similar to~\cite{you2017learning}, as well as over the CLS token $c$ and the tokens of all patches:
\begin{equation}
    \gL(\vx) = \sum_{t\in \gT} \Bigg( \frac{\gL_t(\vx; c) + \frac{1}{|\mathcal{P}|} \sum_{p \in \gP} \gL_t(\vx; p)}{2} \Bigg),
    \label{eq:loss_sum_teachers_cls_patch}
\end{equation}
where $|\mathcal{P}|$ is the number of patch tokens.

\subsection{Protocol summary}
\label{sec:protocol_summary}

We first present a summary of the experimental protocol we use for the analysis in this section. Further details are presented in the \supp.

\paragraph{Datasets and backbones.}
To better isolate the effects of different distillation components, we use the same training data and architectures for all teachers and students, \ie the ImageNet-1K dataset~\cite{russakovsky2015ilsvrc} and ViT-Base~\cite{dosovitskiy2021an}, respectively. During distillation, we discard the labels of ImageNet and only use the images; no supervised loss is combined with
the distillation losses presented above.

\looseness=-1
\paragraph{Teachers.}
We consider models pretrained using \textit{self-supervised learning} (SSL), like DINO~\cite{caron2021emerging} or iBOT~\cite{zhou2022ibot}, and \emph{supervised learning} like DeiT-III~\cite{touvron2022deitiii} or fine-tuned dBoT~\cite{liu2024dbot}, optimized for the classification task of ImageNet-1K.
The former have proven extremely effective for generalization whereas the latter achieve state-of-the-art accuracy on the ImageNet-1K task.
In this section, we present our analysis for $M=2$ teachers, specifically DINO and Deit-III.
More teachers and combinations are explored in~\Cref{sec:performance} and in the \supp.

\paragraph{Tasks.}
We measure performance on many tasks, divided along the following axes:
1)~Top-1 accuracy on the \emph{training set classes} on the ImageNet-1K validation set~\cite{russakovsky2015ilsvrc} (IN-val);
2)~\textit{Transfer learning} performance on unseen classes; we report top-1 accuracy averaged over 15 diverse image classification datasets;\footnote{The 15 datasets are: 5 ImageNet-CoG levels~\cite{sariyildiz2021concept} tailored for concept generalization, 8 small-scale fine-grained datasets
(Aircraft, Cars196, DTD, EuroSAT, Flowers, Pets, Food101, SUN397) and two long-tail datasets (iNaturalist-2018 and 2019).}
\textit{Dense prediction} performance on 3) semantic segmentation and 4) depth estimation; we report mIoU on ADE-20k~\cite{zhou2019semantic} and RMSE on NYUD~\cite{silberman2012indoor}, measured using a protocol that is essentially dense classification, \ie using linear probes as in~\cite{oquab2024dinov2}.
We learn linear probes for all tasks directly over encoder outputs $\vz$.

\subsection{Analyzing multi-teacher distillation of ViT tokens}
\label{sec:tokens}

\looseness=-1
In this section, we analyze and revisit different aspects of distillation that are specific to ViT encoders, \eg the {use} of CLS and patch tokens.
The former is normally fed as input to image-level classifiers while patch tokens are important for dense prediction.
In this section we study their statistics and explore how this affects design choices of the distillation setup.
The top part of~\Cref{tab:analysis} compares the accuracy of the self-supervised DINO and supervised \deit on the different evaluation axes.
They show complementary strengths, \ie they respectively perform
well on transfer learning and the ImageNet-1K validation set (IN-val).

\paragraph{Equalizing feature statistics across tokens and teachers.}
We start by analyzing the statistics of features extracted from the CLS and patch tokens of both teachers and show that this should be taken into account for multi-teacher distillation. We calculate such statistics
and notice
a number of discrepancies in their first and second moment values, both between CLS and patch tokens of a given teacher as well as across teachers. The norm and standard deviation for the CLS token features of DINO, for example, are double the ones for patch tokens of the same model, while the same statistics also differ across \deit and \dino tokens (see \supp for more details).

{
\setlength{\tabcolsep}{4pt}
\setcounter{rownumbers}{0}

\def\colgroupspace{{\hskip 20pt}}
\def\colgroupspacelarge{{\hskip 20pt}}

\begin{table*}[t]
    \centering
    \caption{
        {\textbf{Component analysis for distillation from two teachers.}}
         We report:
        image classification on
        1) ImageNet-1K (IN-val) and
        2) 15 transfer learning datasets (averaged),
        3) semantic segmentation on ADE-20K, and
        4) depth estimation on NYUd.
        Column legend: std: feature standardization, DP: dedicated projector heads for CLS/patch tokens, \dpshort: \dplong and \textit{tdrop}: teacher dropping regularization.
    \label{tab:analysis}
    }
    \adjustbox{max width=\linewidth, center}{
    \begin{tabular}{rl@{\colgroupspace}cccc@{\colgroupspace}cccc}
    \toprule
    & \multirow{2}{*}{Model} & \multirow{2}{*}{std} & \multirow{2}{*}{DP} & \multirow{2}{*}{\dpshort} & \multirow{2}{*}{\textit{tdrop}} &
    IN-val & Transfer & Segmentation & Depth \\
    &&&&&& \footnotesize{top-1 ($\uparrow$)} & \footnotesize{top-1 ($\uparrow$)} & \footnotesize{mIoU ($\uparrow$)} & \footnotesize{RMSE ($\downarrow$)}\\
    \midrule
    \multicolumn{10}{l}{\em \quad Teacher models} \\
    \rownum & \dino &&&&    &  77.7   & 72.4     & 30.4    & 0.570 \\
    \rownum & \deit &&&&   &  83.6 & 68.5     & 32.3    & 0.589 \\
    \rownum & \textit{best teacher} &&&&   &  83.6 & 72.4     & 32.3    & 0.570 \\
    \midrule
    \multicolumn{10}{l}{\em \quad Multi-teacher distillation (DINO \& DeiT-III teachers)} \\
    \rownum & basic setup &&&&&  \degradesmost{78.7} &  \better{73.1} & \good{33.9} & \better{0.560} \\
        \cmidrule{7-10}

    \rownum & \multirow{4}{*}{\bf{\unic}}  & \checkmark &&&&  \degrades{81.4} & \better{73.8} & \better{36.1} & \better{0.558} \\
    \rownum &
    & \checkmark & \checkmark && 	& \degrades{82.2}	& \best{74.1} & \better{36.9}	& \better{0.551} \\
     \rownum &
     & \checkmark & \checkmark & \checkmark &  	& \degradesless{82.7} & \best{74.2} &	\best{37.4} &	\best{0.546} \\

        \rownum &
        & \checkmark & \checkmark & \checkmark &  \checkmark & \ok{83.2} & \better{73.5} & \best{37.3} & \best{0.547} 	\\

    \bottomrule
    \end{tabular}
    }
\end{table*}
}

To explore whether such statistical inconsistencies across features affect distillation, we add feature standardization on each teacher output, \ie we normalize teacher features to zero mean and unit variance before computing the loss, which was shown to be useful in
\cite{heo2019comprehensive}.
This not only equalizes any differences between CLS and patch tokens
but also for tokens across teachers.
For convenience and generality, we propose to learn such normalization statistics on-the-fly during distillation, using an exponential moving average.
From~\Cref{tab:analysis} we see that the performance of models learned via distillation is consistently higher using feature standardization for both image- and patch-level tasks (rows 4 vs. 5).

\insight{Feature standardization improves multi-teacher distillation}

\looseness=-1
\paragraph{Projector heads for CLS and patch tokens.}
Beside statistical differences, the CLS and patch token
are also conceptually different: CLS is a global token expected to encode
image-level semantics whereas the patch tokens encode local information.
To better capture these specifics
from CLS and patch tokens, we experiment with dedicated teacher-specific projector heads for each type of tokens. This comes at no added cost in practice, since we discard the projectors after distillation.
We discuss expendable projectors further in~\Cref{sec:projectors}.
Comparing rows 5 and 6 in~\Cref{tab:analysis} we see that specializing
the teacher-specific projector heads to either CLS {or}
patch tokens leads to further gains.
\insight{Dedicated projectors for CLS/patches improve distillation performance}

\paragraph{Classification on ImageNet and novel classes.}
Results in~\Cref{tab:analysis} show that models learned via multi-teacher distillation lack in terms of ImageNet-1K performance
compared to highly optimized models for that specific task, such as DeiT-III (82.2 vs. 83.6). One may suggest that this is due to the fact that we do not use labels during distillation. To test that, we also performed distillation using \textit{only} the DeiT-III model as a teacher. In that case we were able to reach a top-1 accuracy of 83.1\% on ImageNet. This is much higher than the 82.2\% we get distilling jointly from multiple teachers
and we therefore see that there is still space for improvement during distillation itself.

From
\Cref{tab:analysis} we also see that models learned via multi-teacher distillation greatly outperform DINO on transfer learning and classification of novel classes.
This is also true for the recent \ibot~\cite{zhou2022ibot} model, which also achieves {state-of-the-art} top-1 accuracy, \ie 72.4\% on average for transfer learning on our setup.
\insight{Multi-teacher distillation significantly improves concept generalization}

\looseness=-1
\paragraph{Multi-teacher distillation for dense prediction.}
To assess the discriminative power of patch tokens individually, we consider two dense prediction tasks, semantic segmentation and depth prediction, after linear probing. \Cref{tab:analysis} shows that even the basic multi-teacher distillation setup improves over the best teacher (row 4). More importantly,
performance increases even further (row 6) using standardization and dedicated projectors for the CLS and patch tokens. The student encoder
achieves \textit{+4.6\% higher mIoU} than the best teacher for segmentation. 
This result is even more impressive when compared to the performance of models that are targeting improved dense prediction.
Our models, which are distilled from teachers trained with supervised and contrastive learning
achieve dense prediction performance comparable to models known to excel at dense tasks, \ie models trained via
masked patch prediction like \ibot~\cite{zhou2022ibot}: \ibot achieves 36.6\% mIoU on ADE-20K, while our student reaches 36.9\%.

\insight{Multi-teacher distillation improves the discriminative power of patch tokens}

\paragraph{Retaining complementary teacher strengths.}
From the results in~\Cref{tab:analysis}, we see that models learned with our multi-teacher distillation setup and simple modifications like feature standardization and dedicated projectors for CLS/patch tokens are starting to show strong generalization performance on a number of axes. We will use models distilled under this setup as the basis for the rest of our study. Such models seem to retain the complementary strengths of their teachers: They already outperform the best teacher on transfer learning and dense prediction tasks, while also enjoying decent performance on the ImageNet task.

\insight{Learning from multiple teachers can combine their strengths}

As we discuss above, there is however still room for improvement; we ideally want models to match or outperform the best teacher on all tasks.
In the next
sections, we analyze different aspects of our distillation setup and introduce further improvements towards that end.

\subsection{A ladder of projectors for distillation}
\label{sec:projectors}

The basic setup above uses expendable projector heads as a way of injecting teacher-specific parameters during distillation.\footnote{Projector heads are discarded after distillation and linear probes are learned over the encoder outputs $\vz$.}
Such modules are appended at the end of the encoders and act as small ``buffers'' between the encoder output and the feature space considered by the loss.
In this section, we propose
to use more of these
expendable modules
in a complementary way:
as \textit{information highways}
that propagate information from intermediate layers to the loss in a more direct manner.
Intermediate layers have been
used to improve distillation~\cite{you2017learning,liu2020adaptive,hao2022learning}, typically
by adding extra losses on top of those layers.
However, this leads to a more challenging optimization.
Besides,
hyper-parameter tuning with many added losses
is combinatorial, and it becomes
cumbersome. These issues are far more prominent in the case of multiple teachers.

Instead of adding
losses on intermediate representations, we propose to augment the existing expendable teacher-specific projector head to receive inputs from intermediate layers and append modules that connect all intermediate layer tokens directly to the teacher-specific projector head before the loss. We refer to such augmented projectors as a \emph{\dplong}. This architecture bares similarities to the adaptor architecture that is typically used for adapting a model to a new task~\cite{yin2023parameter}. In our case, however, the adaptor-like modules we append during distillation are \textit{expendable}.

Specifically, we attach MLP projectors to intermediate layers and augment the input of the teacher-specific projectors $h_t$ that until now operated only on the
last layer of the student encoder.
Let $\vz^l$ denote the $l$-th layer output of the student encoder for $l=1,\ldots,L$.
The head for the \dplong{} becomes:

\begin{equation}
    h^{\dpshort}_t(\{\vz^l : l \in L\}) =  \sum_{l = 1}^{L}  h_t^l( \vz^l ),
    \label{eq:dp}
\end{equation}
where $h_t^l$ denotes the MLP projector head attached after layer $l \in L$.
The architecture of $h_t^l$ is identical to $h_t$, however, since we are adding multiple such projector heads, we significantly reduce the hidden dimension $d_h^l$ and set $d_h^l=d$ when $l < L$.
We explore architecture choices in the \supp.

From~\Cref{tab:analysis} we see that this {\dplong} improves performance overall (row 8), especially for dense prediction. It seems that the dense connections lead to better prime patch tokens.
Gains are also significant for
supervised classification:
ImageNet-1K accuracy is increased by +0.5\%.

\insight{A \dplong leads to improvements for both CLS and patch tokens}

\subsection{Learning all teachers equally well}
\label{sec:balancing}

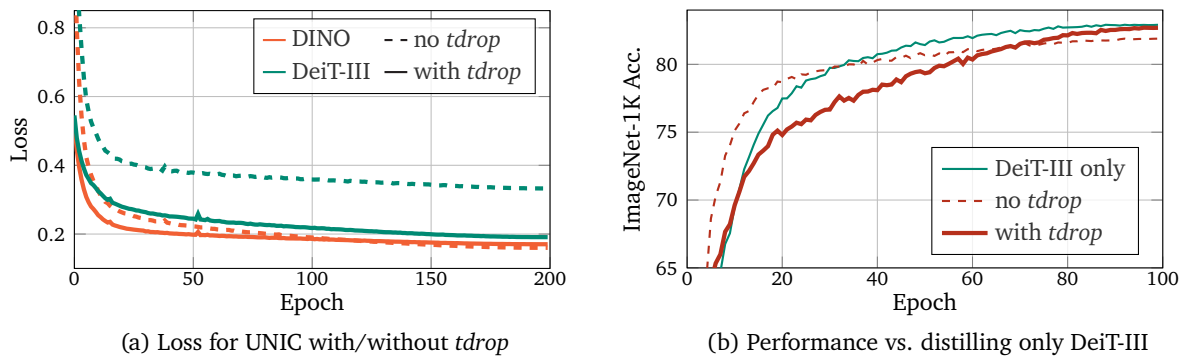
\begin{figure*}[t]
    \centering
    \begin{subfigure}[b]{.49\linewidth}
        \centering
        \begin{tikzpicture}[trim axis right,trim axis left]
\begin{axis}[
    xlabel={Epoch},
    ylabel={Loss},
    ymin=0.1,
    xmin=0,
    xmax=200,
    ymax=0.85,
    width=\linewidth,
    height=\utilheight,
    grid=major,
    label style={font=\small},
    tick label style={font=\scriptsize},
    xlabel shift = -0.15cm,
    ylabel shift = -0.1cm,
    x tick label style={yshift={0.08cm}},
    legend pos=north east,
    legend style={font=\small, legend columns=2},
    legend image post style={scale=0.5},
    legend style={/tikz/every even column/.append style={column sep=0.1cm}}]

\addplot[\dinoc,  thick] coordinates {(83.8, 74.5)};
\addplot[Black,  thick, dashed] coordinates {(83.8, 74.5)};
\addplot[\deitc,  thick] coordinates {(83.8, 74.5)};
\addplot[Black,  thick, ] coordinates {(83.8, 74.5)};

\addplot[\dinoc, ultra thick, dashed] table [col sep=comma, x expr=\coordindex, y index=0] {./tex/res/data_divineb_loss_dino.csv};

\addplot[\deitc, ultra thick, dashed] table [col sep=comma, x expr=\coordindex, y index=0] {./tex/res/data_divineb_loss_deit.csv};

\addplot[\dinoc, ultra thick] table [col sep=comma, x expr=\coordindex, y index=0] {./tex/res/data_divinet_loss_dino.csv};

\addplot[\deitc, ultra thick] table [col sep=comma, x expr=\coordindex, y index=0] {./tex/res/data_divinet_loss_deit.csv};

\legend{\dino, no \emph{tdrop}, \deit, with \emph{tdrop}};

\end{axis}
\end{tikzpicture}
\vspace*{-0.17cm}
        \caption{Loss for \unic with/without \textit{tdrop}}
        \label{fig:tdrop_loss}
    \end{subfigure}
    \hfill
    \begin{subfigure}[b]{.49\linewidth}
        \centering
        \begin{tikzpicture}[trim axis right,trim axis left]
\begin{axis}[
    xlabel={Epoch},
    ylabel={ImageNet-1K Acc.},
    ymin=65, ymax=84,
    label style={font=\small},
    tick label style={font=\scriptsize},
    grid=major,
    xmin=0,
    xmax=100,
    xlabel shift = -0.15cm,
    ylabel shift = -0.1cm,
    x tick label style={yshift={0.08cm}},
    legend style={at={(0.95,0.05)}, anchor=south east, font=\small},
    width=\linewidth,
    height=\utilheight,
]

\addplot[\deitc, thick] table [col sep=comma, x expr=\coordindex, y index=0] {./tex/res/data_lp_cls_deit.csv};

\addplot[\divinec, thick, dashed] table [col sep=comma, x expr=\coordindex, y index=0] {./tex/res/data_lp_cls_divineb.csv};

\addplot[\divinec, ultra thick] table [col sep=comma, x expr=\coordindex, y index=0] {./tex/res/data_lp_cls_divinet.csv};

\addplot[\deitc, thick] coordinates {(83.8, 74.5)};
\addplot[\divinec, thick, dashed] coordinates {(83.8, 74.5)};
\addplot[\divinec, thick, ] coordinates {(83.8, 74.5)};

\legend{\deit only, no \emph{tdrop}, with \emph{tdrop}};

\end{axis}
\end{tikzpicture}
\vspace{-0.17cm}
        \caption{Performance vs. distilling only \deit}
        \label{fig:tdrop_linprobe}
    \end{subfigure}
    \vspace{-0.2cm}
    \caption{\textbf{Analyzing teacher dropping regularization (\textit{tdrop})}.
    {\bf (a)} Loss for each of the two teachers
    during multi-teacher distillation, with and without \textit{tdrop}.
    {\bf (b)} ImageNet-1K top-1 accuracy when distilling from \dino \& \deit together, versus distilling only from \deit, \ie the teacher that excels at this task.
    }
    \label{fig:analysis}
\end{figure*}

The basic setup assumes that the final goal is for
the distilled encoder to represent each teacher equally well.
When distillation uses feature standardization across all teachers and
simple losses like cosine and smooth-$\ell_1$, there exists a straightforward way to compare how much each of the different teachers is learned: One may simply compare the \textit{magnitudes} of the losses, that indicate how well we are approximating the feature space of each teacher.

\looseness=-1
\Cref{fig:tdrop_loss} displays the loss curves for multi-teacher distillation for \unic models, using the setup presented in~\Cref{sec:tokens} (dashed lines). We see that the DINO teacher seems to be learned faster and better than DeiT-III.

\insight{Teachers do not equally contribute without further intervention}

It therefore comes as no surprise that our student lacks performance on ImageNet-1K, \ie the task that DeiT-III excels at.
But what if DINO was not even part of the distillation process? In~\Cref{fig:tdrop_linprobe} we show how ImageNet-1K accuracy changes during distillation
using DINO \& DeiT-III as teachers, and for the case of  distilling \textit{only} from DeiT-III.
We see that our model learns faster using multiple teachers but converges to a lower accuracy:
The student seems to exploit features from the additional teacher to ramp up performance faster,
but fails to reach the accuracy of distilling DeiT-III alone (83.1\%).

\looseness=-1
\Cref{fig:analysis} suggests that some form of loss balancing
could be beneficial.
Loss balancing is common in multi-task settings:
In most cases it is done \textit{manually} by adding hyperparameters that control each loss. Such an approach is however cumbersome for many teachers and losses like our case, something also discussed in~\cite{ranzinger2023radio}. It is important to avoid the combinatorial nature of manual tuning.
Another way, would be to use some of the existing methods for loss balancing that are proposed for multi-task learning, \eg methods like Adaloss~\cite{hu2019learning}.
We argue that the case of multi-teacher distillation over standardized features and simple regression losses is much simpler than multi-task learning when it comes to balancing the losses:
The magnitudes of the losses are comparable and can be used for balancing and pacing the distillation process.

\paragraph{Teacher dropping regularization.}
We introduce a simple
scheme for loss balancing that we name \textit{teacher dropping}. Instead of designing some soft loss weighing algorithm, we take inspiration from methods like randomized dropout~\cite{srivastava2014dropout} and path dropping~\cite{huang2016deep}, and propose to ``drop'', \ie zero-out the loss, for a subset of the teachers.
Dropping teachers at random
is however something that would not encourage loss equalization across teachers. Instead, we propose to directly use absolute magnitudes of the losses when selecting which teachers to drop, \ie keeping the teacher whose loss magnitude is maximal and dropping any other teacher with some probability.
This bares conceptual similarities to adaptive dropout~\cite{ba2013adaptive}, but our method is \textit{non-parametric},
and simply exploits the fact that feature space losses on constrained representations are comparable.

\begin{figure}[t]
    \begin{center}
        \begin{tikzpicture}
\begin{axis}[
    width=\linewidth,
    height=5cm,
    ymin=0,
    ymax=1,
    xlabel={Epoch},
    ylabel={Teacher coeff. ($\alpha_t$)},
    xmin=0,
    xmax=200,
    grid=both,
    legend pos=north east,
    cycle list name=color list,
    label style={font=\small},
    tick label style={font=\scriptsize},
    minor tick num=1,
    legend image post style={scale=0.5},
    legend style={font=\small},
    xlabel shift = -0.15cm,
    ylabel shift = -0.09cm,
    x tick label style={yshift={0.08cm}},
]

\addplot[Black, thick] coordinates {(1, 74.5)};
\addplot[Black, thick, dashed] coordinates {(1, 74.5)};
\addplot[Black, thick, dotted] coordinates {(1, 74.5)};

\addplot[\dinoc, dotted, ultra thick] table [col sep=comma, x expr=\coordindex, y index=0] {./tex/res/data_c_d_dino_p100.csv};

\addplot[\deitc, dotted, ultra thick] table [col sep=comma, x expr=\coordindex, y index=0] {./tex/res/data_c_d_deit_p100.csv};

\addplot[\dinoc, dashed, ultra thick] table [col sep=comma, x expr=\coordindex, y index=0] {./tex/res/data_c_d_dino_p050.csv};

\addplot[\deitc, dashed, ultra thick] table [col sep=comma, x expr=\coordindex, y index=0] {./tex/res/data_c_d_deit_p050.csv};

\addplot[\dinoc, ultra thick] table [col sep=comma, x expr=\coordindex, y index=0] {./tex/res/data_c_d_dino_p025.csv};

\addplot[\deitc, ultra thick] table [col sep=comma, x expr=\coordindex, y index=0] {./tex/res/data_c_d_deit_p025.csv};

\legend{$p=0.25$, $p=0.5$, $p=1.0$};

\end{axis}
\end{tikzpicture}
\vspace*{-0.15cm}
    \end{center}
    \caption{
        {\bf Teacher coefficients} $\alpha_t$ during distillation from {\color{\deitc}DeiT} and {\color{\dinoc}\dino}.
    }
    \label{fig:teacher_coeff}
\end{figure}
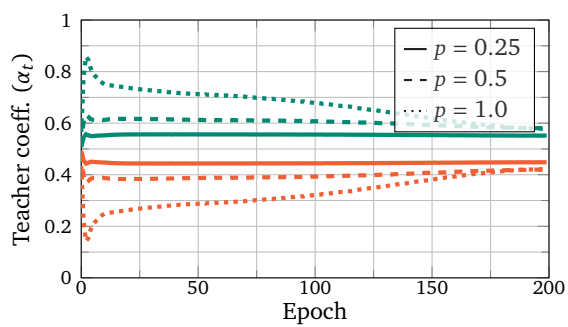

We perform loss-based teacher dropping at the image level.
At each iteration and for every image, we define a binary coefficient $\alpha_t = \{0,1\}$ for each teacher $t$ that is multiplied with the corresponding loss $\gL_t$.
This determines whether teacher $t$ would be dropped or not for that image with probability $p$.
To make sure there is always
some signal to learn from,
we choose to never drop the teacher with the maximum magnitude loss, \ie the teacher that the current model approximates least well.
All other teachers could be dropped with probability $p$.
Specifically and for each image, the coefficient for teacher $t \in \gT$ is given by:
\begin{equation}
{\small
  \alpha_t =
    \begin{cases}
      1 & \text{if $\gL_t = \max_t \gL_t$}, \\
      (1 - \delta) & \text{if $\gL_t  \neq \max_i \gL_i$, with $\delta \sim $ Bernoulli$(p)$  }.\\
    \end{cases}
    }
    \label{eq:teacher_dropping_alpha}
\end{equation}
In all cases, the teacher that is least well approximated in the current iteration will always be used.
We also experimented with patch-level teacher dropping but found no noticeable gains (see \supp).

{
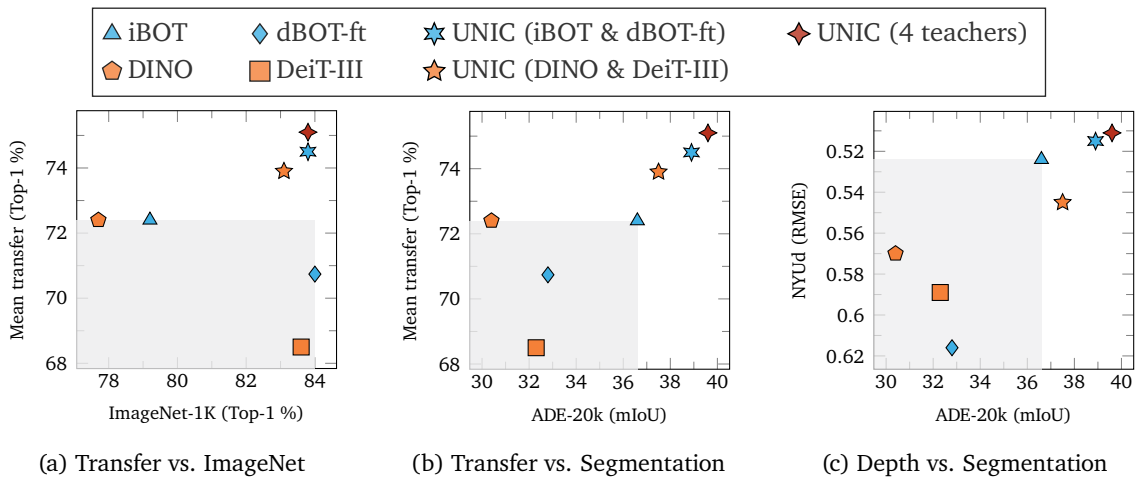
\begin{figure*}[th!]
    \centering
    \adjustbox{width=0.8\linewidth}{
        \begin{tikzpicture}
            \begin{axis}[
            hide axis,
            xmin=10,
            xmax=50,
            ymin=0,
            ymax=0.4,
            legend style={draw=white!15!black,legend cell align=left, font=\footnotesize},
            legend columns=4,
            legend style={/tikz/every even column/.append style={column sep=15pt}}
            ]
            \addlegendimage{ibot}
            \addlegendentry{\ibot};
            \addlegendimage{dbot}
            \addlegendentry{\dbot};
            \addlegendimage{bdivine}
            \addlegendentry{{\unic} (\ibot\& \dbot)};
            \addlegendimage{fdivine}
            \addlegendentry{{\unic} (4 teachers)};
            \addlegendimage{dino}
            \addlegendentry{\dino};
            \addlegendimage{deit}
            \addlegendentry{\deit};
            \addlegendimage{ddivine}
            \addlegendentry{{\unic} (\dino\& \deit)};
            \end{axis}
        \end{tikzpicture}
    }
    \\
    \vspace{4pt}
    \adjustbox{max width=\linewidth}{
    \begin{subfigure}[t]{.32\textwidth}
        \centering
        \begin{tikzpicture}[inner frame sep=0pt]
    \begin{axis}[
        height=\plothightwidth,
        width=\plothightwidth,
        xlabel={ImageNet-1K (Top-1 \%)},
        ylabel={Mean transfer (Top-1 \%)},
        label style={font=\tiny},
        tick label style={font=\tiny},
        minor tick num=1,
        xtick={},ytick={},
        legend pos=south west,
        x tick label style={yshift={0.05cm}},
        inner frame sep=0pt,
        label style={font=\scriptsize},
        tick label style={font=\scriptsize},
        legend style={font=\scriptsize},
    ]
        \fill[opacity=0.8,fill=\teacherareac] (0,0) rectangle (84,72.4);

        \addplot[dino] coordinates {(77.7, 72.4)};
        \addplot[deit] coordinates {(83.6, 68.5)};
        \addplot[dbot] coordinates {(84.0, 70.74)};
        \addplot[ibot] coordinates {(79.2,72.4)};

        \addplot[ddivine] coordinates {(83.1, 73.9)};

        \addplot[bdivine] coordinates {(83.8, 74.5)};

        \addplot[fdivine] coordinates {(83.8, 75.1)};

    \end{axis}
\end{tikzpicture}
        \caption{Transfer vs. ImageNet}
        \label{fig:2d_plots_gen_im1k}
    \end{subfigure}
    \hfill
    \begin{subfigure}[t]{.32\textwidth}
        \centering
        \begin{tikzpicture}[inner frame sep=0pt]
    \begin{axis}[
        height=\plothightwidth,
        width=\plothightwidth,
        xlabel={ADE-20k (mIoU)},
        ylabel={Mean transfer (Top-1 \%)},
        label style={font=\tiny},
        tick label style={font=\tiny},
        minor tick num=1,
        legend pos=south east,
        legend style={font=\tiny},
        x tick label style={yshift={0.05cm}},
        label style={font=\scriptsize},
        tick label style={font=\scriptsize},
        legend style={font=\scriptsize},
    ]

        \addplot[dino] coordinates {(30.4, 72.4)};
        \addplot[deit] coordinates {(32.3, 68.5)};
        \addplot[dbot] coordinates {(32.8, 70.74)};
        \addplot[ibot] coordinates {(36.6,72.4)};

        \fill[opacity=0.8,fill=\teacherareac] (0,0) rectangle (36.6,72.4);

        \addplot[ddivine] coordinates {(37.5, 73.9)};

        \addplot[bdivine] coordinates {(38.9, 74.5)};

        \addplot[fdivine] coordinates {(39.6, 75.1)};

    \end{axis}
\end{tikzpicture}
        \caption{Transfer vs. Segmentation}
        \label{fig:2d_plots_gen_seg}
    \end{subfigure}
    \hfill
    \begin{subfigure}[t]{.32\textwidth}
        \centering
        \begin{tikzpicture}[inner frame sep=0pt]
    \begin{axis}[
        height=\plothightwidth,
        width=\plothightwidth,
        xlabel={ADE-20k (mIoU)},
        ylabel={NYUd (RMSE)},
        label style={font=\tiny},
        tick label style={font=\tiny},
        minor tick num=1,
        y dir=reverse,
        legend pos=south east,
        legend style={font=\tiny},
         x tick label style={yshift={0.05cm}},
        inner frame sep=0pt,
        label style={font=\scriptsize},
        tick label style={font=\scriptsize},
        legend style={font=\scriptsize},
    ]

        \fill[opacity=0.7,fill=\teacherareac] (0, 0.7) rectangle (36.6, 0.524);

        \addplot[dino] coordinates {(30.4, 0.570)};
        \addplot[deit] coordinates {(32.3, 0.589)};
        \addplot[dbot] coordinates {(32.8, 0.616)};
        \addplot[ibot] coordinates {(36.6, 0.524)};
        \addplot[ddivine] coordinates {(37.5, 0.545)};
        \addplot[bdivine] coordinates {(38.9, 0.515)};
        \addplot[fdivine] coordinates {(39.6, 0.511)};

    \end{axis}
\end{tikzpicture}
        \caption{Depth vs. Segmentation}
        \label{fig:2d_plots_depth_seg}
    \end{subfigure}
    }
    \caption{
        {
        {\bf Performance of different \unic encoders on different pairs of tasks.}
        We report performance for \unic encoders distilled from \dino \& \deit, \ibot \& \dbot and distilling from all four teachers together.
        We show results on ImageNet-1K {\bf (a)}, over 15 transfer learning tasks {\bf (a, b)}, semantic segmentation {\bf (b, c)} and depth estimation {\bf (c)}.
        }
    }
    \label{fig:2d_plots}
\end{figure*}
}

\paragraph{Effect of teacher dropping during distillation.}
We study the impact of
teacher dropping during distillation in~\Cref{fig:tdrop_loss}:
teacher dropping makes the loss magnitudes of the teachers
much more similar
as training progresses (solid lines).
In~\Cref{fig:teacher_coeff} we plot how the teacher coefficients $\alpha_t$ vary during distillation;
teacher utilization becomes more balanced and stabilizes after some epochs.

\insight{Teachers are distilled equally well with teacher dropping regularization}

\paragraph{How does teacher dropping affect performance?}
We compared teacher dropping regularization to manually balancing the teacher losses,
{random dropping~\cite{fukuda2017efficientkd}}, as well as to the recent Adaloss~\cite{hu2019learning} loss balancing method. Starting from results in row 6 in~\Cref{tab:analysis},
we found that none of these strategies is able to noticeably improve, let alone outperform results with teacher dropping (row 8).
Specifically, Adaloss achieves 80.1/73.6/34.3/0.565 on the four tasks, respectively (see \supp for details).
Besides performance, we believe the effectiveness and simplicity of the proposed teacher dropping is unparalleled.

We studied the impact of the teacher dropping probability $p$ and found performances to be stable for different values.
Yet, a higher $p$ favours ImageNet performance, with a slight decrease on tasks where the student already outperforms the best teacher (see \supp).

From~\Cref{tab:analysis} (row 8) we see that teacher dropping boosts performance for ImageNet-1K, \ie improves distillation on the task where our distilled models were lacking the most. When combining teacher dropping with a \dplong, we are able to achieve 83.2\%, our top performance on that task.
This performance is only 0.4\% lower than the highly optimized \deit (row 3).
What is more, we have also closed the observed gap between multi-teacher distillation and specialized distillation using \deit alone.
Teacher dropping significantly contributes to that end, increasing performance by 0.5\% over our best model with \dplong (rows 7 vs. 8).

\insight{Teacher dropping regularization is a simple and effective way to balance teachers, specifically designed for multi-teacher distillation}

\subsection{Towards universal classification models}

\looseness=-1
Multi-teacher distillation using a \dplong and teacher dropping regularization enables us to reach ImageNet classification performance comparable to the highly optimized \deit,
while simultaneously outperforming the best teacher on transfer learning performance on 15 datasets with mostly novel classes including long-tail ones, as well as on patch-level classification tasks like semantic segmentation and depth estimation.
We contend this evidence demonstrates that
our distilled models operate as more \textit{universal} classification models.
We will refer to models learned with our enhanced multi-teacher distillation setup
as \textbf{\unic} models
(which stands for \underline{UNI}versal \underline{C}lassification, pronounced ``\textit{unique}'').

\section{Experimental study}
\label{sec:performance}

\paragraph{Teachers.}
In~\Cref{sec:results} we report our main results distilling from two pairs of teachers (\deit~\cite{touvron2022deitiii} \& \dino~\cite{caron2021emerging} and \ibot~\cite{zhou2022ibot} \& \dbot~\cite{liu2024dbot}\footnote{We use the dBOT model fine-tuned on ImageNet-1K.}),
as well as using all four together.
In all cases we use publicly available ViT-Base/16 models trained on ImageNet-1K.
In~\Cref{sec:arbitrary} we further present results when distilling larger teachers that are trained on arbitrary data.

\paragraph{Extended protocol.}
We use the protocol summarized in~\Cref{sec:protocol_summary} and detailed in the \supp.
We additionally report results on ImageNet-v2~\cite{recht2019imagenet}, an alternative validation set for ImageNet, as well as two datasets for measuring performance under domain shift, \ie ImageNet-R~\cite{hendrycks2021many} and ImageNet-Sketch~\cite{wang2019learning}.
Besides reporting results for all 15 transfer datasets jointly, we further split the datasets into separate axes, \ie for concept generalization~\cite{sariyildiz2021concept}, long-tail~\cite{van2018inaturalist} and small-scale fine-grained recognition datasets
(Aircraft~\cite{maji2013aircraft}, Cars196~\cite{krause2013cars}, DTD~\cite{cimpoi2014texture}, EuroSAT~\cite{helber2019eurosat}, Flowers~\cite{nilsback2008flowers}, Pets~\cite{parkhi2012cats}, Food101~\cite{bossard2014food101}, SUN397~\cite{xiao2010sun}).

In all cases we chose hyperparameters based on ImageNet-1K performance, the task which corresponds to the distillation data.
See the \supp for further implementation and evaluation details. There, we further report results using the pre-existing classifiers in a plug-and-play manner, as well as for the case of distillation using synthetic data from the ImageNet-SD dataset~\cite{sariyildiz2023fake}.

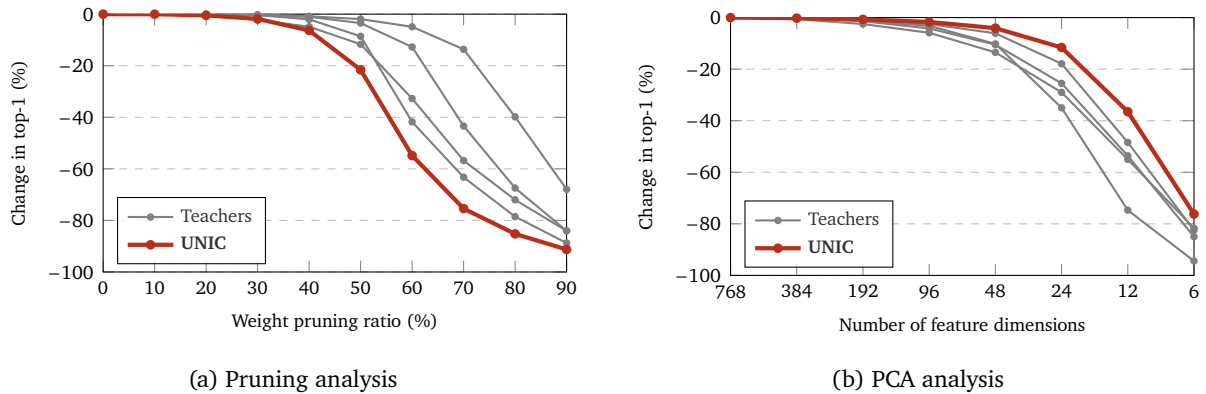
\begin{figure*}[t]
    \centering
    \begin{subfigure}[t]{.48\linewidth}
        \centering
        \begin{tikzpicture}
\begin{axis}[
    width=\linewidth,
    height=\utilheight,
    xlabel={Weight pruning ratio (\%)},
    ylabel={Change in top-1 (\%)},
    ylabel style={align=center},
    xtick={0,1,2,3,4,5,6,7,8,9},
    xticklabels={0,10,20,30,40,50,60,70,80,90},
    legend pos=south west,
    xmin=0,
    xmax=9,
    ymin=-100,
    ymax=0,
    ymajorgrids=true,
    grid style=dashed,
    label style={font=\tiny},
    tick label style={font=\tiny},
    legend style={font=\tiny},
    label style={font=\scriptsize},
    tick label style={font=\scriptsize},
    legend style={font=\scriptsize},
]

\addplot[teachutil, thick] coordinates {
    (0,0.00) (1,0.00) (2,0.00) (3,-0.39) (4,-1.03) (5,-3.47) (6,-12.74) (7,-43.37) (8,-67.44) (9,-84.04)
};

\addplot[teachutil, thick] coordinates {
    (0,0.00) (1,0.00) (2,-0.12) (3,-0.24) (4,-0.84) (5,-1.91) (6,-4.90) (7,-13.64) (8,-39.83) (9,-67.94)
};

\addplot[teachutil, thick] coordinates {
    (0,0.00) (1,0.00) (2,0.00) (3,-0.38) (4,-2.02) (5,-8.59) (6,-41.79) (7,-63.26) (8,-78.54) (9,-88.76)
};

\addplot[teachutil, thick] coordinates {
    (0,0.00) (1,0.00) (2,-0.36) (3,-2.26) (4,-4.88) (5,-11.67) (6,-32.74) (7,-56.79) (8,-72.02) (9,-84.05)
};

\addplot[fdivineline, ultra thick] coordinates {
    (0,0.00) (1,-0.02) (2,-0.47) (3,-1.83) (4,-6.32) (5,-21.59) (6,-54.84) (7,-75.41) (8,-85.23) (9,-91.28)
};

\legend{Teachers,,,,\bf{\unic}};

\end{axis}
\end{tikzpicture}
        \vspace{-10pt}
        \caption{
            {Pruning analysis}
        }
        \label{fig:pruning_in1k}
    \end{subfigure}
    \hfill
    \begin{subfigure}[t]{.48\linewidth}
        \centering
        \begin{tikzpicture}
\begin{axis}[
    width=\linewidth,
    height=\utilheight,
    xlabel={Number of feature dimensions},
    ylabel={Change in top-1 (\%)},
    ylabel style={align=center},
    xmin=0,
    xmax=7,
    ymin=-100,
    ymax=0,
    xtick={0,1,2,3,4,5,6,7},
    xticklabels={768,384,192,96,48,24,12,6,3},
    legend pos=south west,
    ymajorgrids=true,
    grid style=dashed,
    label style={font=\tiny},
    tick label style={font=\tiny},
    legend style={font=\tiny},
    label style={font=\scriptsize},
    tick label style={font=\scriptsize},
    legend style={font=\scriptsize},
]

\addplot[teachutil, thick] coordinates {
    (0,0.0) (1,-0.6) (2,-2.5) (3,-5.9) (4,-13.5) (5,-29.0) (6,-55.0) (7,-81.7)
};

\addplot[teachutil, thick] coordinates {
    (0,0.0) (1,0.0) (2,-1.4) (3,-4.2) (4,-10.5) (5,-25.5) (6,-53.6) (7,-85.0)
};

\addplot[teachutil, thick] coordinates {
    (0,0.0) (1,-0.4) (2,-1.0) (3,-2.5) (4,-6.1) (5,-17.9) (6,-48.4) (7,-82.5)
};

\addplot[teachutil, thick] coordinates {
    (0,0.0) (1,-0.4) (2,-1.2) (3,-3.2) (4,-10.2) (5,-35.0) (6,-74.7) (7,-94.4)
};

\addplot[fdivineline, ultra thick] coordinates {
    (0,0.0) (1,-0.2) (2,-0.7) (3,-1.7) (4,-4.1) (5,-11.6) (6,-36.5) (7,-76.2)
};

\legend{Teachers,,,,\bf{\unic}};

\end{axis}
\end{tikzpicture}
        \vspace{-10pt}
        \caption{
            {PCA analysis}
        }
        \label{fig:pca_in1k}
    \end{subfigure}
    \vspace{-5pt}
    \caption{
        {\bf Network utility analysis} via ImageNet-1K linear probing for the four teachers
        and our
        student \unic distilled from all of them.
        For each model, before training linear probes, we either {\bf (a)} prune their weights or {\bf (b)} reduce the dimension of their features via PCA.
        We report change in top-1 accuracy compared to their base performance.
        \unic's encoder weights work together more cohesively {\bf (a)}, and its feature space is more robust to dimensionality reduction {\bf (b)}.
    }
    \label{fig:network_utility}
\end{figure*}

\subsection{Results}
\label{sec:results}
We summarize results for our best \unic models from different teachers in \Cref{fig:spider,fig:2d_plots}.
In~\Cref{fig:spider} we show \textit{relative} gains for a \unic model trained from all four teachers,
while in \Cref{fig:2d_plots} we report results for models distilled from three different sets of teachers (\dino \& \deit, \ibot \& \dbot and all four teachers).
A short summary of our most important observations follows.
\begin{enumerate}
\setlist[enumerate]{itemsep=0mm}
\item \textbf{Stronger teachers give stronger students.}
From \Cref{fig:2d_plots} we see that \ibot \& \dbot yield improved student models
compared to \dino \& \deit.
\item \textbf{Adding more teachers seems to generally improve performance.}
Distilling from all four teachers produces an even stronger student for most cases.
This is also true when the additional teachers are not better than the existing ones:
Besides ImageNet and transfer, adding \dino \& \deit to the ensemble also improves segmentation performance over \ibot.
\item \textbf{\unic models excel at image-level classification}.
\unic from 4 teachers attains $83.8\%$ and $80.3\%$ top-1 accuracy on ImageNet-1K and ImageNet-v2, matching the top performance of the state-of-the-art \dbot model ($84\%$ and $80\%$, respectively).
Results are also strong on transfer learning, with \unic achieving $+2.7\%$ higher top-1 on average than \ibot/\dino.
\item \textbf{Impressive gains on transfer to small fine-grained datasets}. \unic achieves a \textit{$+9.2\%$ relative gain} on average on 8 small-scale classification datasets, some for domains far outside the ImageNet training set used for all teachers and distillation (\ie including satellite images and textures).
Complementary teachers appear to be highly beneficial in this case.
\item \textbf{Strong gains for dense prediction with linear probing}. Strong gains are also observed on segmentation and depth estimation, for example on ADE-20K where \unic achieves a $+8.2\%$ relative gain over \ibot. Although far from being the optimal protocol for the task, linear probing is best to evaluate the discriminative power of the patch tokens from the encoder.
\item \textbf{Retaining top teacher performance for domain shifts}.
\deit shows exceptionally high performance on ImageNet-R and Sketch ($51.4\%$ and $39.3\%$ top-1 accuracy, respectively).
Our best \unic model retains this top performance, achieving $51.4\%$ and $38.5\%$, respectively.
\end{enumerate}

\begin{table*}
    \begin{center}
    \caption{
        {\bf Results after distilling MetaCLIP-Huge/14 and DINOv2-Giant/14} into a ViT-Large student.
            The UNIC and AM-RADIO~\cite{ranzinger2023radio} models are distilled using the ImageNet-1K dataset.
            Results for teachers and AM-RADIO are from~\cite{ranzinger2023radio}.
            Note that AM-RADIO uses the DINOv2 model with registers as a teacher which achieves slightly higher performance for semantic segmentation (reported below).
        }
    \adjustbox{max width=\linewidth}{
    \begin{tabular}{lccc}
    \toprule
    \multirow{2}{*}{Model} & $k$-NN & Zero-shot   & ADE-20K \\
                           & top-1 acc. & top-1 acc. &  mIoU \\
    \midrule
    \multicolumn{4}{l}{\em Teacher Models} \\
    ~MetaCLIP-Huge/14~\cite{xu2023metaclip}               & 82.1           & 80.5          & 35.4          \\
    ~DINOv2-Giant/14-reg~\cite{oquab2024dinov2}              & 83.4           &     --        & {\bf 48.7} \\ \midrule
    AM-RADIO~\cite{ranzinger2023radio} & 84.8           & 80.4          & 48.1          \\
    \textbf{UNIC}-L & {\bf 85.6} & {\bf 81.4} & 48.3         \\
    \bottomrule
    \end{tabular}
    }
    \label{tab:amradio}
    \end{center}
\end{table*}

\subsection{Weight and feature space utilization}
\label{sec:utilization}

In this section, we seek to better understand why multi-teacher distillation leads to overall stronger encoders.
We do that by investigating the utilization of the encoder weights after pruning (\Cref{fig:pruning_in1k}) and the feature space after dimensionality reduction (\Cref{fig:pca_in1k}).
We report the change in accuracy on ImageNet-1K for our \unic model and
its
teachers when we prune the weights or reduce the feature dimension before training linear probes.
We prune encoder weights using $\ell_1$-norm-based unstructured weight pruning, and perform dimensionality reduction
using PCA with whitening.

From~\Cref{fig:pruning_in1k}, we see that the performance of \unic drops more rapidly than any of the teachers as we increase the pruning ratio.
This indicates that the encoder weights show improved synergy, working together more cohesively and efficiently to enhance the model's overall performance.

\insight{\unic encoders utilize weights more effectively}

At the same time, in~\Cref{fig:pca_in1k}, we see that our student preserves its base performance better than all teachers as we reduce the number of dimensions with PCA.
It seems that the feature space of \unic can be represented better with fewer principal components, possibly because of higher entanglement in the original feature space.

\insight{\unic encoders are more resilient to dimensionality reduction}

\subsection{Distilling arbitrary models}
\label{sec:arbitrary}

In this section, we extend our study to larger teachers trained on arbitrary datasets, \ie MetaCLIP ViT-Huge/14~\cite{xu2023metaclip} and DINOv2 ViT-Giant/14~\cite{oquab2024dinov2}.
We distill a ViT-Large/14 student from these two teachers, initially at resolution 224  for 200 epochs and then at resolution 336 for 100 additional epochs.
We set the teacher dropping probability $p$ to $0.25$.
In~\Cref{tab:amradio} we report the UNIC model performance for $k$-NN and zero-shot classification on ImageNet-1K, as well as semantic segmentation on ADE-20K.
We further compare to the recent AM-RADIO~\cite{ranzinger2023radio}, an approach that resembles our base setup with dedicated projectors.
These results offer some basic verification that our insights are also valid in this more generic distillation case: Using teacher dropping regularization and a ladder of projectors enables  UNIC models to outperform both teachers in the majority of cases.

\section{Conclusions}
\label{sec:conc}

In this paper,
we systematically analyze
multi-teacher distillation and introduce improvements to the distillation process
that significantly enhance the performance of student models across various benchmarks.
More importantly, we show that it is possible to distil from multiple teachers with complementary strengths and learn models that match or improve the respective best teacher in both image- and patch-based classification tasks.
In that regard, we view \unic models as \textit{universal} classification models,
{advancing the frontier of general representation learning without task-specific adaptation.}

\paragraph{Acknowledgements.}
The authors would like to sincerely thank Myung-Ho Ju, Florent Perronnin, Rafael Sampaio de Rezende, Vassilina Nikoulina and Jean-Marc Andreoli for inspiring discussions and many thoughtful comments.

{
    \small
    \bibliographystyle{ieeenat_fullname}
    \bibliography{main}
}

\setcounter{table}{0}
\setcounter{figure}{0}
\renewcommand\thefigure{\Alph{figure}}
\renewcommand\thetable{\Alph{table}}

\etocdepthtag.toc{mtappendix}
\etocsettagdepth{mtappendix}{subsection}
\etocsettagdepth{mtchapter}{none}

\appendix
\section*{Appendix}

{
In {this} \supp, we present implementation details (\textbf{\Cref{sec:supp_implementation}}) as well as further details on the different evaluation protocols we use (\textbf{\Cref{sec:supp_protocol}}). We also present additional analysis and experiments, specifically:
    \begin{itemize}
        \item \textbf{\Cref{sec:supp_extended_results}} presents extended results for three different teacher combinations as well as results studying the impact of teacher dropping and the \dplong independently of each other for all scenarios.
        \item \textbf{\Cref{sec:supp_plug_and_play}} presents results when using \unic models with pre-existing classifiers (plug-and-play).
        \item \textbf{\Cref{sec:imagenet_sd}} presents results for distillation using only synthetic images from ImageNet-SD~\cite{sariyildiz2023fake} dataset.
        \item \textbf{\Cref{sec:student_arch}} presents results when we distill the four teachers into a ViT-Small architecture.
        \item \textbf{\Cref{sec:token_statistics}} details feature statistics on the CLS and patch tokens for the four teachers.
        \item \textbf{\Cref{sec:supp_projectors,sec:supp_tdrop}} present ablations regarding the expandable projectors and teacher dropping, respectively.
        \item \textbf{\Cref{sec:supp_utilization}} presents an analysis on the utilization of weights and features for the task of semantic segmentation.
    \end{itemize}
}
\section{Implementation details}
\label{sec:supp_implementation}

\paragraph{Data.}
We train all models on ImageNet-1K~\cite{russakovsky2015ilsvrc} without using image labels.

\paragraph{Distillation resolution and data augmentation.}
For data augmentation we use random resized crop to produce $224\times224$ images, then apply random horizontal flip, color jitter, grayscale, Gaussian blur, and solarization, mostly following~\cite{touvron2022deitiii}.

\paragraph{Models.}
Unless otherwise explicitly stated, all teachers and student models have the same encoder architecture: a ViT-Base~\cite{dosovitskiy2021an} with patch size 16.
For teachers, we download the official weights for their encoders from the authors' repositories.\footnote{Code repositories for the teacher models:\\
    \scriptsize
    \dino: \url{https://github.com/facebookresearch/dino} \\
    \deit: \url{https://github.com/facebookresearch/deit} \\
    \ibot: \url{https://github.com/bytedance/ibot} \\
    \dbot: \url{https://github.com/liuxingbin/dbot} \\
    DINOv2: \url{https://github.com/facebookresearch/dinov2} \\
    MetaCLIP: \url{https://github.com/facebookresearch/MetaCLIP}
}
$d_h$ and $d_h^l$ for \dplong{} (\dpshort) are set to 3072 and 768, respectively.
For teacher dropping regularization (\tdrop), we use image-level dropping with a probability of $0.25$ when distilling from two teachers, and $0.5$ when distilling from four teachers.

\paragraph{Optimization.}
Unless otherwise stated, models are trained for 100 epochs.
When we use teacher dropping regularization and drop teacher losses, we train for longer, \ie 200 epochs.
It is worth noting that training the base model for double the epochs only shows small improvements.

As for the distillation loss, we minimize the combination of cosine and smooth-$\ell_1$ losses between the outputs of student ($\vs$) and teacher ($\vt$):
\begin{align}
    \gL^{cos} (\vs, \vt) &= 1 - \frac{\vs \cdot \vt}{||\vs||_2 \times ||\vt||_2}, \\
    \gL^{sl1} (\vs, \vt) &=
\begin{cases}
    0.5 \times ||\vs - \vt||_2^2, & \text{for } ||\vs - \vt||_1 < 1, \\
    ||\vs - \vt||_1 - 0.5, & \text{otherwise}.
\end{cases}
\end{align}

We use the AdamW~\cite{loshchilov2019decoupled} optimizer, with a learning rate of $3\mathrm{e}{-4}$, weight decay of $3\mathrm{e}{-2}$, batch size of 512 split across 4 GPUs.
We apply a linear warmup for the learning rate during the first 10 epochs, then decrease it with a cosine schedule~\cite{loshchilov2017sgdr}.

\paragraph{Details for distilling arbitrary models.}
In~\Cref{tab:amradio}, we distill DINOv2-G/14 and MetaCLIP-H/16 into a ViT-Large/14 student in two stages, first at resolution 224 for 200 epochs, following our normal setup, and then we further fine-tune the model at resolution 336 for 100 more epochs.
Since the feature dimensions of both student and teacher are higher, we increased hidden dimension of \dpshort{}:
$d_h$ and $d_h^l$ are set to 4096 and 1024, respectively.

\paragraph{Zero-shot classification experiment.}
In our experiment using arbitrary models (DINOv2 and MetaCLIP) as teachers (\Cref{tab:amradio}), we evaluate our UNIC model's performance on zero-shot classification.
Since the feature space dimensionality of UNIC is different from the features output by the MetaCLIP text encoder, we further used the projector of the MetaCLIP teacher during inference as a way of making the feature spaces compatible.
This was \textit{the only} experiment where we did not utilize the UNIC encoder features directly.

{
\setlength{\tabcolsep}{8pt}
\begin{table*}[h!]
\centering
\caption{
    {
    {\bf Distillation from different teacher combinations.}
    We report results on four task axes for different distillation setups and teacher combinations: Distilling from a single teacher (rows 5-8), distillation from \dino \& \deit (rows 9-12), from \ibot \& \dbot (rows 13-16), and from all four teachers (rows 17-20).
    We report results for the strong ``Base setup'', \ie our basic distillation setup enhanced with feature standardization and dedicated projector heads for CLS/patch tokens (row 6 of Tab.~1 from the main paper) as well as when using the proposed ladder of projectors (LP) and teacher dropping regularization (\textit{\tdrop}) separately on top of the base setup.
    Finally, we report performance using both LP and \textit{tdrop} (\unic models). The best performance over each column among the methods in each group is bolded.
    }
    \vspace{-0.2cm}
}
\label{tab:supp_extended_results}
\adjustbox{max width=\linewidth}{
\begin{tabular}{llcccc}
\toprule
& \multirow{2}{*}{Method} & IN-val & Transfer & Segmentation & Depth \\
& & \footnotesize{top-1 ($\uparrow$)} & \footnotesize{top-1 ($\uparrow$)} & \footnotesize{mIoU ($\uparrow$)} & \footnotesize{RMSE ($\downarrow$}) \\
\midrule
\rowcolor{tabdefault}
\multicolumn{6}{l}{\em Teachers} \\
\rownumft & \dino & 77.7 & {\bf 72.4} & 30.4 & 0.570 \\
\rownumft & \ibot & 79.2 & {\bf 72.4} & {\bf 36.6} & {\bf 0.524} \\
\rownumft & \deit & 83.6 & 68.5 & 32.3 & 0.589 \\
\rownumft & \dbot & {\bf 84.0} & 70.7 & 32.8 & 0.616 \\
\midrule
\rowcolor{tabdefault}
\multicolumn{6}{l}{\em Distillation from a single teacher} \\
\rownumft & \dino & 77.3 & {\bf 72.9} & 31.2 & 0.568 \\
\rownumft & \deit & 83.1 & 71.6 & 35.4 & 0.571 \\
\rownumft & \ibot & 79.0 & {\bf 72.9} & {\bf 36.9} & {\bf 0.531} \\
\rownumft & \dbot & {\bf 83.4} & 72.3 & 35.9 & 0.563 \\
\midrule
\rowcolor{tabdefault}
\multicolumn{6}{l}{\em Distillation from \dino \& \deit} \\
\rownumft & {Base} setup & 82.2 & 74.1 & 36.9 & 0.551 \\
\rownumft & \quad + LP & 82.7 & {\bf 74.2} & 37.4 & 0.546 \\
\rownumft & \quad + \textit{tdrop} (no LP) & 83.0 & 74.0 & 36.7 & 0.553  \\
\rownumft & \unic & {\bf 83.1} & 73.9 & {\bf 37.5} & {\bf 0.545} \\
\midrule
\rowcolor{tabdefault}
\multicolumn{6}{l}{\em Distillation from \ibot \& \dbot} \\
\rownumft & {Base} setup & 82.7 & 74.4 & 39.1 & 0.518 \\
\rownumft & \quad + LP & 83.2 & {\bf 74.8} & {\bf 39.7} & {\bf 0.505} \\
\rownumft & \quad + \textit{tdrop} (no LP) & 83.5 & 74.3 & 38.4 & 0.525 \\
\rownumft & \unic & {\bf 83.8} & 74.5 & 38.9 & 0.515 \\
\midrule
\rowcolor{tabdefault}
\multicolumn{6}{l}{\em Distillation from all four teachers} \\
\rownumft & {Base} setup & 82.8 & 74.5 & 38.5 & 0.539 \\
\rownumft & \quad + LP & 83.3 & {\bf 75.1} & {\bf 39.7} & 0.518  \\
\rownumft & \quad  + \textit{tdrop} (no LP)  & 83.6 & 74.7 & 38.5 & 0.522 \\
\rownumft & \unic & {\bf 83.8} & {\bf 75.1} & {39.6} & {\bf 0.511} \\
\bottomrule

\end{tabular}
}
\end{table*}
}

\section{Further details on the evaluation \\ protocols}
\label{sec:supp_protocol}

We perform a range of downstream tasks to evaluate the performance of models, including
image classification on ImageNet-1K~\cite{russakovsky2015ilsvrc} and 15 transfer datasets,
semantic segmentation on ADE-20K~\cite{zhou2019semantic}, and
depth estimation on NYUd~\cite{silberman2012indoor}.

\paragraph{Image-level classification tasks.}
We measure performance on the ImageNet-1K validation set~\cite{russakovsky2015ilsvrc}, on ImageNet-v2~\cite{recht2019imagenet}, an alternative validation set for ImageNet, as well as on two datasets for measuring performance under domain shift, \ie ImageNet-R~\cite{hendrycks2021many} and ImageNet-Sketch~\cite{wang2019learning}.

We measure transfer learning performance on 15 datasets:
5 ImageNet-CoG levels~\cite{sariyildiz2021concept} tailored for concept generalization,
8 small-scale fine-grained datasets
(Aircraft~\cite{maji2013aircraft}, Cars196~\cite{krause2013cars}, DTD~\cite{cimpoi2014texture}, EuroSAT~\cite{helber2019eurosat}, Flowers~\cite{nilsback2008flowers}, Pets~\cite{parkhi2012cats}, Food101~\cite{bossard2014food101}, SUN397~\cite{xiao2010sun}), and
two long-tail datasets (iNaturalist-2018 and 2019~\cite{van2018inaturalist}).

{All tasks are formulated as classification tasks using  linear probes attached directly to frozen encoder outputs $\vz$.}
Each linear probe is trained separately for each {dataset.}
{We} follow~\cite{sariyildiz2023trex} and train linear logistic regression classifiers on top of encoder outputs.
For all models (both teachers and students), we extract features from the CLS token, except for \dbot, which does not include a CLS token.
Following the original implementation of \dbot~\cite{liu2024dbot}, we extract the global average pooling (GAP) features instead.
We then train a linear classifier using pre-extracted features, \ie \textit{we do not use data augmentation at this stage}.
This is the reason why we report slightly lower performance on the ImageNet-1K validation set for our teacher models via this approach, {\ie
compared to the performances reported in the respective papers.}
For fairness, we follow this process also for all models (including teachers and students), so that linear probing setups are identical in both cases.
Hyper-parameters for the linear classifiers are tuned using Optuna~\cite{optuna2019} and scikit-learn~\cite{scikitlearn}.
For all image classification results, we use at test time the resolution used during distillation.

\paragraph{Dense prediction tasks.}
Semantic segmentation and depth estimation are dense prediction tasks, both formulated as classification tasks in this work, and solved following the {simple} setup proposed in \cite{oquab2024dinov2}.
It uses features from patch tokens, extracted from the last output layer of the frozen {encoder} and used as input to a linear prediction head.
For semantic segmentation, the linear head is trained to predict class logits from a patch token. This yields a $32 \times 32$ logit map, {which is} further upsampled {via bilinear interpolation}
to {the} resolution {of} $512\times512$ to obtain a segmentation map.

{For depth estimation, the features extracted from the last layer of the frozen {encoder} are first upsampled {via bilinear interpolation} by a factor {of} $4$, then concatenated along the feature dimension with the CLS token, and {finally} used as input to a linear layer.}
Depth prediction is treated as a soft classification task using AdaBins~\cite{adabins} with 256 uniformly distributed bins.

\paragraph{Reporting a performance summary over all tasks.}
As metrics vary across tasks (\ie top-1 accuracy for classification, mIoU for segmentation and RMSE for depth estimation), in Fig.~1 of the main paper we report \textit{relative} performance for each task, which is calculated
on each task as the difference between the performance of our \unic model distilled from four teachers to that of the best teacher, divided by that same best performance.

\section{Extended analysis and results}
\label{sec:supp_experiments}

\subsection{{Extended results and component \\ ablations}}
\label{sec:supp_extended_results}

\looseness=-1
In~\Cref{tab:supp_extended_results}, we report results when distilling from two sets of teachers, as well as distilling from all four.
{We report results for a number of distillation configurations: a) a ``base setup'', which is our basic distillation setup detailed in Section 3.1 of the main paper, plus feature standardization and dedicated projectors for CLS and patch tokens; a very strong baseline to beat, b) using a ladder of projectors (LP) over the base setup, c) using teacher dropping (\tdrop) over the base setup and d) results for \unic models, \ie models trained using the base setup plus a ladder of projectors and teacher dropping regularization.

\looseness=-1
We see that both LP and \tdrop show improved gains, with LP maximizing the gains for dense prediction tasks, but still lacking on ImageNet-1K, the task most complementary to the rest for the selected teachers. When using \tdrop without LP, we see that it can achieve strong balance over the tasks that the teachers are complementary at, but dense prediction performance is not really improved. When using both modifications together, we see that we get the best possible results overall, with ImageNet-1K performance now reaching the performance of the best teacher.
}

For completeness, we report in~\Cref{tab:spider} all the results used to generate~\Cref{fig:spider}.

\begin{table*}
    \begin{center}
    \caption{\textbf{Relative gains using our \unic} encoder distilled from four teachers (\dino, \deit, \ibot, \dbot), over the respective best teacher for each task. UNIC solves all classification tasks using a \textit{single encoder} and no task-specific parameters. \textit{DS} refers to domain shift datasets (ImageNet-R~\cite{hendrycks2021many} and ImageNet-Sketch~\cite{wang2019learning}), \textit{CoG} to the 5 ImageNet-CoG levels~\cite{sariyildiz2021concept}, \textit{LT} to two long-tail datasets (iNaturalist-2018 and 2019~\cite{van2018inaturalist}) and \textit{FG} to the 8 small-scale fine-grained datasets
(Aircraft~\cite{maji2013aircraft}, Cars196~\cite{krause2013cars}, DTD~\cite{cimpoi2014texture}, EuroSAT~\cite{helber2019eurosat}, Flowers~\cite{nilsback2008flowers}, Pets~\cite{parkhi2012cats}, Food101~\cite{bossard2014food101}, SUN397~\cite{xiao2010sun}).}
    \adjustbox{max width=\linewidth}{
    \begin{tabular}{lcccccccc}
        \toprule

Model & IN-1K & IN-V2& DS & CoG & LT & FG & ADE20K & NYUd \\
\midrule
~\textit{Teachers} \\
\dino   & 77.7 & 74.0 & 32.6 & 65.3 & 81.6 & 53.0 & 30.4 & 0.57 \\
\deit   & 83.6 & 79.6 & 45.3 & 64.0 & 77.3 & 44.8 & 32.3 & 0.59 \\
\ibot   & 79.2 & 75.3 & 33.3 & 65.9 & 81.4 & 52.4 & 36.6 & 0.52 \\
\dbot   & 84.0 & 80.0 & 44.5 & 65.8 & 78.8 & 50.7 & 32.8 & 0.61 \\
\midrule
\textbf{UNIC} & 83.8 & 80.3 & 45.0 & 68.2 & 83.7 & 57.9 & 39.6 & 0.51 \\ ~~\textit{rel. gains} &
    \loss{-0.2}& \gain{0.4}& \loss{0.6} & \gain{3.5}& \gain{2.6}& \gain{9.2} & \gain{8.2} & \gain{2.4}\\
\bottomrule
    \end{tabular}
    }
    \label{tab:spider}
    \end{center}
\end{table*}

\looseness=-1
\paragraph{Distilling from a single teacher.}
In~\Cref{tab:supp_extended_results}, we also show results after using our distillation setup to distil from each teacher independently.
By simply using a form of \textit{self-distillation} we see that the transfer learning performance of \deit and \dbot, the two models tuned for ImageNet-1K, increases significantly.
One explanation is that since the
features at the output of the student encoder are followed by a projector, they might have become more generic than the ones from teachers, which are tailored for the task.
We see similar but smaller gains on that axis also for the self-supervised models \dino and \ibot.

\paragraph{Effect of fine-tuning at a higher dimension for UNIC-L.} In~\Cref{tab:amradio} we present results for larger UNIC models, \ie using a ViT-L student. These models are first distilled at resolution 224 for 200 epochs, following our normal setup, and then further fine-tuned at resolution 336 for 100 more epochs.
In~\Cref{tab:amradio_supp_extended} we report performance before and after the fine-tuning step.

\begin{table*}
    \begin{center}

    \caption{
        {\bf Effect of finetuning at a higher resolution.} When distilling MetaCLIP-Huge/14 and DINOv2-Giant/14 into a ViT-Large student (UNIC-L), we first distill the model from scratch for 200 epochs at resolution 224 and then fine-tune for 100 more epochs at resolution 336. Results after each phase of training are presened below.
        For all UNIC models we set teacher dropping probability $p$ to $0.25$. UNIC models denoted with $^*$ do not use a ladder of projectors.
        }
    \adjustbox{max width=\linewidth}{
    \begin{tabular}{lccccc}
    \toprule
    \multirow{2}{*}{Model}         & \multirow{2}{*}{Epochs}     &       \multirow{2}{*}{Resolution}              & $k$-NN & Zero-shot   & ADE-20K\\
    &&&top-1 acc.& top-1 acc. &  mIoU\\
    \midrule
    \multirow{2}{*}{\textbf{UNIC}-L$^*$} &  200 & 224     & 85.0 & 80.7 &   47.7 \\
    &  ~~+100 & 336 & 85.1 & 81.1 & \textbf{49.1} \\
    \midrule
    \multirow{2}{*}{\textbf{UNIC}-L}     &  200 & 224     & 85.4 & 81.2 & 47.1         \\
    &  ~~+100 & 336 & {\bf 85.6} & {\bf 81.4} & 48.3         \\
    \bottomrule
    \end{tabular}
    }
    \label{tab:amradio_supp_extended}
    \end{center}
\end{table*}

{
\setlength{\tabcolsep}{4pt}
\begin{table}[t]
\centering
\caption{
    {\bf Plug-and-play} performance on the ImageNet-1K validation set.
    For our \unic models distilled from either one of the teacher pairs or all four of them, we report their logistic regression (LogReg) and plug-and-play evaluations using the pre-existing classifiers from the best supervised teacher (DeiT-III for the first row which reaches 83.5 top-1 accuracy, dBOT-ft for the second and third rows, which reaches 84.5 top-1 accuracy).
    For LogReg (which is our default evaluation protocol for image classification tasks in this paper), we train linear logistic regression classifiers on top of pre-extracted encoder representations.
    For plug-and-play, we use the pre-existing ImageNet-1K classifiers
    {from the teacher which are fed from the projected student features; this does not require any task-specific training for the student.}
}
\label{tab:supp_plug_and_play}
\begin{tabular}{lcc}
\toprule
Model & LogReg & Plug-and-play \\
\midrule
\unic (DINO \& DeiT-III) & 83.1 & 83.3 \\
\unic (iBOT \& dBOT-ft) & 83.6 & 83.8 \\
\unic (4 teachers) & 83.8 & 84.0 \\
\bottomrule
\end{tabular}
\end{table}
}

\subsection{Results with pre-existing classifiers (plug-and-play)}
\label{sec:supp_plug_and_play}

The student is trained together with teacher-specific projector(s) that mimic the teacher features. It is thus possible to directly use a task head, learned with teacher features, and directly plug it on top of the corresponding teacher projectors we learn together with the student encoder.
\Cref{tab:supp_plug_and_play} shows the results on the ImageNet-1K validation set when using the pre-existing classifiers from the public \deit and \dbot models as well as using linear probes trained with our protocol.

\looseness=-1
We see that the \emph{plug-and-play} scenario {can} lead to better results {using the projectors rather than the original student features}.
This shows that heads trained for a specific teacher can be directly used without any retraining.
The higher accuracies can also be explained by the fact that our evaluation protocol does not include data augmentation for efficiency reasons (see~\Cref{sec:supp_protocol}), or that the projectors add extra parameters on top of the encoder.

{
\setlength{\tabcolsep}{8pt}
\begin{table*}[t!]
\centering
\caption{\textbf{Multi-teacher distillation using synthetic data.} We replace ImageNet-1K with ImageNet-SD~\cite{sariyildiz2023fake} for distilling UNIC models. ImageNet-SD is an ImageNet-sized dataset composed of synthetic images generated
with
Stable Diffusion~\cite{rombach2022high} using the ImageNet class prompts; we refer the reader to~\cite{sariyildiz2023fake} for more details.}
\label{tab:supp_sd}
\adjustbox{max width=\linewidth}{
\begin{tabular}{llcccc}
\toprule
& \multirow{2}{*}{Method} & IN-val & Transfer & Segmentation & Depth \\
& & \footnotesize{top-1 ($\uparrow$)} & \footnotesize{top-1 ($\uparrow$)} & \footnotesize{mIoU ($\uparrow$)} & \footnotesize{RMSE ($\downarrow$}) \\
\midrule
\rowcolor{tabdefault}
\multicolumn{6}{l}{\em Teachers (Trained on ImageNet-1K)} \\
\rownumft & \dino & 77.7 & 72.4 & 30.4 & 0.570 \\
\rownumft & \ibot & 79.2 & 72.4 &  36.6 & 0.524 \\
\rownumft & \deit & 83.6 & 68.5 & 32.3 & 0.589 \\
\rownumft & \dbot & \textbf{84.0} & 70.7 & 32.8 & 0.616 \\
\midrule
\rowcolor{tabdefault}
\multicolumn{6}{l}{\em Multi-teacher distillation using \textbf{ImageNet-1K} or \imsd{ImageNet-1K-SD}~\cite{sariyildiz2023fake}} \\
\rownumft & \textbf{\unic} & {\bf 83.8} & {\bf 75.1} & {\bf 39.6} & {\bf 0.511} \\
\rownumft & \textbf{\unic}\imsd{-SD} & 81.7 & {\bf 74.7} & {\bf 37.8} & {\bf 0.528} \\
\bottomrule

\end{tabular}
}
\end{table*}
}

\subsection{Distilling using synthetic images from ImageNet-SD}
\label{sec:imagenet_sd}

In a recent study, Sariyildiz \etal~\cite{sariyildiz2023fake} replace the ImageNet-1K dataset for supervised training with \emph{ImageNet-SD}, an ImageNet clone composed of Stable Diffusion~\cite{rombach2022high} images obtained using the ImageNet class names as prompts.

In~\Cref{tab:supp_sd} we report results when using this dataset for distillation instead of ImageNet-1K. We see that the UNIC model distilled exclusively on synthetic images is outperforming the best teacher on transfer learning and semantic segmentation. Similar to the observations in~\cite{sariyildiz2023fake}, we also see that performance on classifying the dataset classes decreases. The decrease is however relatively small: the student is better than teachers like iBot or DINO, and outperformed only by the teacher optimized for this specific classification task.

\subsection{{Distilling into a ViT-Small student}}
\label{sec:student_arch}

{
\setlength{\tabcolsep}{8pt}
\begin{table*}[t!]
\centering
\caption{
{{\bf Distilling four ViT-Base/16 teachers into different student architectures.}}
The ``Num. Params.'' column refers to the number of trainable parameters in the encoder of the student architecture.
}
\label{tab:arch}
\adjustbox{max width=\linewidth}{
\begin{tabular}{lcccccc}
\toprule
\multirow{2}{*}{Method} & Student & Num. &   IN-val & Transfer & Segmentation & Depth \\
& Architecture & Params. & \footnotesize{top-1 ($\uparrow$)} & \footnotesize{top-1 ($\uparrow$)} & \footnotesize{mIoU ($\uparrow$)} & \footnotesize{RMSE ($\downarrow$}) \\
\midrule
\rowcolor{tabdefault}
\textbf{\unic} & ViT-Base/16 & 85.8M & 83.8 & 75.1 & 39.6 & 0.511 \\
\textbf{\unic} & ViT-Small/16 & 21.7M & 81.4 & 71.6 & 36.1 & 0.564 \\
\bottomrule
\end{tabular}
}
\end{table*}
}

\looseness=-1
In~\Cref{tab:arch}, we report results when distilling the four teachers into a smaller student architecture, ViT-Small/16.
{Our ViT-Small UNIC model also matches the performance of a ViT-Small \deit on ImageNet~1K}.\footnote{See \url{https://github.com/facebookresearch/deit/blob/main/README_revenge.md}}

\subsection{Statistics for CLS and patch tokens}
\label{sec:token_statistics}

In~\Cref{tab:supp_norm} we report norm and standard deviation for CLS and patch token features from all our teacher models, computed on the ImageNet-1K validation set. We see large variations in the moments, not only across teachers but also across CLS and patch tokens of the same model.

{
\setlength{\tabcolsep}{12pt}
\newcommand{\tabcelltwol}[2]{\begin{tabular}[c]{@{}c@{}}#1\\ #2\end{tabular}}

\begin{table*}[t]
    \centering
    \caption{
        {\bf Feature statistics} obtained on the the ImageNet-1K validation set.
        For each teacher, we extract their encoder outputs, as we do in our evaluations.
        ``CLS'' refers to features of the CLS token, while
        ``Patch'' refers to  patch token features, where the statistics are computed after global average pooling (GAP) applied spatially.
        ``Avg. norm per sample'' (resp. ``Avg. std per sample'') is the average $\ell_2$ norm (resp. standard deviation) of features computed over samples.
        ``Avg. std per dimension'' is the average standard deviation computed over dimensions.
        \dbot does not contain a CLS token.
        When we distill from \dbot, we use its GAP features.
        \vspace{-0.2cm}
    }
    \label{tab:supp_norm}
    \adjustbox{max width=\textwidth}{
    \begin{tabular}{@{}lcccc@{}}
    \toprule
    Model & \tabcelltwol{Feature}{Type} & \tabcelltwol{Avg. norm}{per sample} & \tabcelltwol{Avg. std}{per sample} & \tabcelltwol{Avg. std}{per dimension} \\
    \midrule
    \dino  & CLS & 66.6 & 2.4 & 2.2 \\
    \deit  & CLS & 23.3 & 0.8 & 0.5 \\
    \ibot  & CLS & 69.9 & 2.5 & 2.3 \\
    \midrule
    \dino  & {Patch} & 31.3 & 1.1 & 0.5 \\
    \deit  & {Patch} & 26.2 & 0.9 & 0.5 \\
    \ibot  & {Patch} & 36.3 & 1.3 & 0.9 \\
    \dbot  & {Patch} & 9.8 & 0.4 & 0.4 \\
    \bottomrule
    \end{tabular}
    }
\end{table*}
}

\subsection{Expendable projector ablations}
\label{sec:supp_projectors}

\paragraph{Top-only projector heads.}
We employ such projector heads when not using the \dplong.
In~\Cref{tab:proj_ablations}, we vary the number of hidden
layers in top-only projector heads when distilling from \dino and \deit, and check how they impact performance across all tasks.
Hidden ($d_h$) and output layer dimensions are set to 3072 and 768, similar to the original ViT-Base specification~\cite{dosovitskiy2021an}.
We see that having 1 hidden and output layers (which is highlighted in gray) is the best for ImageNet-1K classification and NYUd depth estimation.

{
\setlength{\tabcolsep}{4pt}
\begin{table*}[t]
\centering
\caption{
    {\bf Architecture of the student projector} used in the absence of the ladder of projectors.
    Results are reported for distillation from \dino and \deit without using \tdrop but using feature standardization and dedicated projectors.
    We vary the number of hidden and output layers in the projectors.
    Number of units for hidden and output layers are 3072 and 768, respectively.
    The row corresponding to the default setup in our experiments is colored in light gray.
    \vspace{-0.2cm}
}
\label{tab:proj_ablations}
\begin{tabular}{cccccc}
\toprule
\multicolumn{2}{c}{Projector} & IN-val & Transfer & Segmentation & Depth \\
Hidden L. & Output L. & \footnotesize{top-1 ($\uparrow$)} & \footnotesize{top-1 ($\uparrow$)} & \footnotesize{mIoU ($\uparrow$)} & \footnotesize{RMSE ($\downarrow$)} \\
\midrule
-- & -- & 81.1 & 73.0 & 34.1 & 0.564 \\
-- & 1  & 81.5 & 73.1 & 35.3 & 0.567 \\
\rowcolor{tabdefault}
1  & 1  & 82.2 & 74.1 & 36.9 & 0.551 \\
2  & 1  & 81.8 & 74.2 & 36.9 & 0.559 \\
3  & 1  & 81.1 & 74.2 & 37.0 & 0.559 \\
\bottomrule
\end{tabular}
\end{table*}
}

{

\begin{table*}[h!]
    \centering
    \caption{
        \textbf{Architecture for the ladder of projector.}
        We vary the hidden dimension of the non-final block (768 by default) as well as which intermediate blocks are connected in the ladder (by default, all, \ie $\{1,..,11\}$).
        Results are reported for distillation from \dino and \deit without using {\em tdrop} but using feature standardization and dedicated projectors.
        The row corresponding to the default setup in our experiments is colored in light gray.
    \vspace{-0.2cm}
    }
    \label{tab:lp_ablations}
    \adjustbox{max width=\linewidth, center}{
    \begin{tabular}{rccccc}
    \toprule
    Hidden & \multirow{2}{*}{Blocks} &
    IN-val & Transfer & Segmentation & Depth \\
    dim. && \footnotesize{top-1 ($\uparrow$)} & \footnotesize{top-1 ($\uparrow$)} & \footnotesize{mIoU ($\uparrow$)} & \footnotesize{RMSE ($\downarrow$)} \\
    \midrule
    64 & $\{1,..,11\}$ & 81.9 & 74.5 & 36.1 & 0.549 \\
    192 & $\{1,..,11\}$ & 82.3 & 74.5 & 36.9 & 0.540 \\
    384 & $\{1,..,11\}$ & 82.5 & 74.4 & 37.8 & 0.547 \\
    \rowcolor{tabdefault}
    768 & $\{1,..,11\}$ & 82.7 & 74.2 & 37.4 & 0.546 \\
    1536 & $\{1,..,11\}$ & 82.7 & 74.5 & 37.7 & 0.544 \\
    \midrule
    768 & $\{6\}$ & 82.0 & 74.6 & 36.7 & 0.545 \\
    768 & $\{3,6,9\}$ & 82.3 & 74.3 & 37.3 & 0.545 \\
    768 & $\{9,10,11\}$ & 82.0 & 74.4 & 37.1 & 0.542 \\
    768 & $\{2,4,6,8,10\}$ & 82.5 & 74.4 & 37.8 & 0.541 \\
    \bottomrule
    \end{tabular}
    }
\end{table*}
}

\paragraph{\CapitalizeFirst{\dplong}.}
{When using} the ladder of projectors, features from intermediate blocks of the student encoder are
projected with a teacher-specific MLP and
summed together with the
{outputs of the projector attached to the last encoder layer}.
In~\Cref{tab:lp_ablations}, we ablate the number of hidden dimensions $d_h^l$ in the MLPs of intermediate blocks, as well as which intermediate blocks are considered. Regarding the hidden dimensions, we see that
performance improves for ImageNet-1K
{as the hidden dimension increases},
{up to a} plateau after 384 for semantic segmentation. To keep the number of parameters relatively small, we thus chose 768. Regarding which blocks to consider, the impact is overall limited as long as sufficient blocks are considered, and considering all of them lead to the best performance on ImageNet-1K.

\subsection{Teacher dropping ablations}
\label{sec:supp_tdrop}

\paragraph{Impact of \tdrop granularity and probability.}
In~\Cref{tab:tdrop_ablation}, we study the impact of the teacher dropping probability $p$ on performance, when \tdrop is used  with and without \dpshort{} and varying the dropping probability between $0$ and $1$.
We see that increasing the dropping probability (\ie training with sparser teachers) leads to generally better performance on ImageNet-1K, while, lower probability leads to better performance on the remaining of the tasks (for transfer learning, semantic segmentation and depth estimation).
Specifically, higher dropping probability $p$ improves performance on the tasks where the ``underlearned'' teacher excells, \ie DeiT-III and ImageNet for the case of DINO and DeiT-III teachers. One can therefore adjust $p$ according to the desired performance on the tasks of the teacher(s) with generally higher loss.

{In the same table, we further study the impact that \tdrop granularity has, \ie when dropping losses on the image or patch level, with the former being the default in all our experiments.
We see no noticeable gains when dropping teachers at the patch level}.

\begin{figure*}[!h]
    \centering
    \begin{subfigure}[t]{.48\linewidth}
        \centering
        \begin{tikzpicture}
\begin{axis}[
    width=\linewidth,
    height=\utilheight,
    xlabel={Weight pruning ratio (\%)},
    ylabel={Segmentation (mIoU$\uparrow$}),
    ylabel style={align=center},
    xtick={0,1,2,3,4,5,6,7,8,9},
    xticklabels={0,10,20,30,40,50,60,70,80,90},
    legend pos=south west,
    xmin=0,
    xmax=9,
    ymin=0,
    ymax=40,
    ymajorgrids=true,
    grid style=dashed,
    label style={font=\tiny},
    tick label style={font=\tiny},
    legend style={font=\tiny},
    label style={font=\scriptsize},
    tick label style={font=\scriptsize},
    legend style={font=\scriptsize},
]

\addplot[teachutil, thick] coordinates {
(0,30.44) (1,29.98) (2,29.96) (3,30.30) (4,29.51) (5,27.46) (6,21.41) (7,13.33) (8,8.02) (9,4.04)
};

\addplot[teachutil, thick] coordinates {
(0,32.28) (1,31.88) (2,31.28) (3,30.28) (4,27.59) (5,23.47) (6,19.11) (7,13.31) (8,7.48) (9,2.60)
};

\addplot[teachutil, thick] coordinates {
(0,36.56) (1,36.53) (2,36.38) (3,35.68) (4,33.56) (5,28.01) (6,15.40) (7,9.03) (8,4.96) (9,2.86)
};

\addplot[teachutil, thick] coordinates {
(0,32.81) (1,32.69) (2,31.74) (3,29.77) (4,27.62) (5,23.74) (6,16.18) (7,10.33) (8,7.09) (9,4.52)
};

\addplot[fdivineline, ultra thick] coordinates {
(0,39.57) (1,39.32) (2,38.81) (3,37.14) (4,32.59) (5,22.28) (6,12.40) (7,6.29) (8,4.21) (9,3.34)
};

\legend{Teachers,,,,\bf{\unic}};

\end{axis}
\end{tikzpicture}
        \caption{
            {Pruning analysis}
        }
        \label{fig:pruning_semseg}
    \end{subfigure}
    \hfill
    \begin{subfigure}[t]{.48\linewidth}
        \centering
        \begin{tikzpicture}
\begin{axis}[
    width=\linewidth,
    height=\utilheight,
    xlabel={Number of feature dimensions},
    ylabel={Segmentation (mIoU$\uparrow$)},
    ylabel style={align=center},
    xmin=0,
    xmax=7,
    ymin=0,
    ymax=40,
    xtick={0,1,2,3,4,5,6,7},
    xticklabels={768,384,192,96,48,24,12,6,3},
    legend pos=south west,
    ymajorgrids=true,
    grid style=dashed,
    label style={font=\tiny},
    tick label style={font=\tiny},
    legend style={font=\tiny},
    label style={font=\scriptsize},
    tick label style={font=\scriptsize},
    legend style={font=\scriptsize},
]

\addplot[teachutil, thick] coordinates {
(0,30.25) (1,24.48) (2,19.95) (3,16.02) (4,11.06) (5,6.36) (6,3.12) (7,1.69)
};

\addplot[teachutil, thick] coordinates {
(0,31.95) (1,28.74) (2,25.63) (3,20.87) (4,13.62) (5,5.95) (6,1.85) (7,0.63)
};

\addplot[teachutil, thick] coordinates {
(0,36.17) (1,34.77) (2,31.43) (3,27.10) (4,21.40) (5,13.18) (6,5.02) (7,2.09)
};

\addplot[teachutil, thick] coordinates {
(0,32.78) (1,29.39) (2,25.14) (3,17.84) (4,8.78) (5,3.05) (6,1.28) (7,0.59)
};

\addplot[fdivineline, ultra thick] coordinates {
(0,39.42) (1,37.45) (2,35.17) (3,31.29) (4,24.18) (5,14.81) (6,6.36) (7,3.07)
};

\legend{Teachers,,,,\bf{\unic}};

\end{axis}
\end{tikzpicture}
        \caption{
            {PCA analysis}
        }
        \label{fig:pca_semseg}
    \end{subfigure}
    \vspace{-5pt}
    \caption{
        {\bf Network utility analysis for semantic segmentation} linear probing for the four teachers
        and our student \unic distilled from all of them.
        For each model, before training linear probes, we either {\bf (a)} prune their weights or {\bf (b)} reduce the dimension of their features via PCA.
        We report the mIoU scores on ADE-20K.
        \unic's encoder weights work together more cohesively {\bf (a)}, and its feature space is more robust to dimensionality reduction {\bf (b)}.
    }
    \label{fig:network_utility_semseg}
    \vspace{-0.5cm}
\end{figure*}
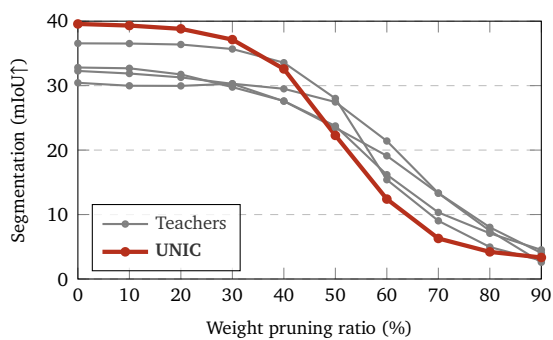
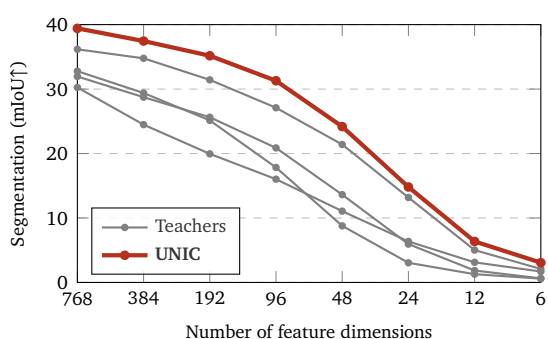

{
\setlength{\tabcolsep}{4pt}
\begin{table*}[!h]
\centering
\caption{
    {\bf Impact of {\em tdrop} {probability and granularity}.}
    We vary the probability between 0 and 1, and the granularity to be either at the image or patch level.
    We show results for distillation from \ibot \& \dbot, without using a ladder of projectors.
    We use feature standardization and dedicated projectors in all cases.
    \vspace{-0.2cm}
}
\label{tab:tdrop_ablation}
\adjustbox{max width=\linewidth}{
\begin{tabular}{ccccccc}
\toprule
\multicolumn{2}{c}{\em tdrop} & LP & IN-val & Transfer & Segmentation & Depth \\
gran. & prob. && \footnotesize{top-1 ($\uparrow$)} & \footnotesize{top-1 ($\uparrow$)} & \footnotesize{mIoU ($\uparrow$)} & \footnotesize{RMSE ($\downarrow$)} \\
\midrule
Image & 0.00 & -- & 83.0 & 74.4 & 39.1 & 0.518 \\
Image & 0.25 & -- & 83.1 & 74.3 & 38.7 & 0.522 \\
Image & 0.50 & -- & 83.5 & 74.3 & 38.4 & 0.525 \\
Image & 1.00 & -- & 83.5 & 73.9 & 37.9 & 0.530 \\
\midrule
Patch & 0.50 & -- & 83.2 & 74.3 & 38.7 & 0.532 \\
Patch & 1.00 & -- & 83.3 & 74.1 & 38.0 & 0.533 \\
\midrule
Image & 0.00 & \checkmark & 83.2 & 74.8 & 39.7 & 0.505 \\
Image & 0.25 & \checkmark & 83.6 & 74.5 & 39.4 & 0.506 \\
Image & 0.50 & \checkmark & 83.8 & 74.5 & 38.9 & 0.515 \\
Image & 1.00 & \checkmark & 83.7 & 73.6 & 38.1 & 0.530 \\
\bottomrule
\end{tabular}
}
\end{table*}
}

\paragraph{Comparing teacher dropping regularization to alternatives.}
In~\Cref{tab:loss_balancing}, we compare \tdrop to AdaLoss~\cite{hu2019learning}, another automatic loss balancing technique, and manual balancing of losses when distilling from all four teachers.
For manual balancing, it is computationally demanding to find the optimal teacher weights due to its combinatorial nature.
We choose 5 different intuitive combinations to see the relative impact of each teacher.
We see that \tdrop achieves significantly better performance than AdaLoss on ImageNet-1K and segmentation, while being comparable to AdaLoss on the remaining tasks.
In the case of manual balancing, no single combination leads to best performance on all tasks.
{
\setlength{\tabcolsep}{3pt}
\begin{table*}[!h]
\centering
\caption{
    {\bf Loss balancing techniques} for distillation from all four teachers (\dino, \deit, \ibot and \dbot).
    We use feature standardization and dedicated projectors in all cases.
    The best {({resp.} second best)} performance over each column among the methods in each group is bolded {({resp.} underlined).
    All experiments performed over the base setup, \ie using feature standardization and dedicated projectors for CLS/patch tokens and without using a ladder of projector heads.}
    \vspace{-0.2cm}
}
\label{tab:loss_balancing}
\adjustbox{max width=\linewidth}{
\begin{tabular}{lcccc}
\toprule
& IN-val & Transfer & Segmentation & Depth \\
Method & \footnotesize{top-1 ($\uparrow$)} & \footnotesize{top-1 ($\uparrow$)} & \footnotesize{mIoU ($\uparrow$)} & \footnotesize{RMSE ($\downarrow$}) \\
\midrule
\multicolumn{5}{l}{\em Manual balancing} \\
{\color{gray} \dino{}$\times$}{\bf 1} + {\color{gray} \deit{}$\times$}{\bf 1} + {\color{gray} \ibot{}$\times$}{\bf 1} + {\color{gray} \dbot{}$\times$}{\bf 1} & 82.2 & {\bf 74.5} & \underline{38.5} & 0.539 \\
{\color{gray} \dino{}$\times$}{\bf 4} + {\color{gray} \deit{}$\times$}{\bf 1} + {\color{gray} \ibot{}$\times$}{\bf 1} + {\color{gray} \dbot{}$\times$}{\bf 1} & 80.6 & 74.0 & 36.1 & 0.549 \\
{\color{gray} \dino{}$\times$}{\bf 1} + {\color{gray} \deit{}$\times$}{\bf 4} + {\color{gray} \ibot{}$\times$}{\bf 1} + {\color{gray} \dbot{}$\times$}{\bf 1} & 83.2 & 74.0 & 37.4 & 0.548 \\
{\color{gray} \dino{}$\times$}{\bf 1} + {\color{gray} \deit{}$\times$}{\bf 1} + {\color{gray} \ibot{}$\times$}{\bf 4} + {\color{gray} \dbot{}$\times$}{\bf 1} & 81.1 & 74.1 & 38.2 & \underline{0.533} \\
{\color{gray} \dino{}$\times$}{\bf 1} + {\color{gray} \deit{}$\times$}{\bf 1} + {\color{gray} \ibot{}$\times$}{\bf 1} + {\color{gray} \dbot{}$\times$}{\bf 4} & {\bf 83.5} & 74.2 & 38.4 & {\bf 0.532} \\
\midrule
\multicolumn{5}{l}{\em Automatic balancing} \\
AdaLoss & 81.9 & {\bf 74.5} & 38.4 & 0.536 \\
{Teacher dropping} {(\tdrop)}
& \underline{83.1} & \underline{74.4} & {\bf 38.8} & \underline{0.533} \\
\bottomrule
\end{tabular}
}
\vspace{0.2cm}
\end{table*}
}

\subsection{Extended results on weight and feature space utilization}
\label{sec:supp_utilization}
In Section 4.2 of the main paper, we study the network utility for teachers and our best \unic model in terms of the utility of their weights and CLS features for ImageNet-1K classification.
We extend this analysis for semantic segmentation, this time, using patch tokens.
From the results shown in~\Cref{fig:network_utility_semseg}, we see that our observations from the main paper are consistent.
When varying the weight pruning ratio (\Cref{fig:pruning_semseg}), \unic's performance drops significantly faster than the ones from the teachers,
meaning that the weights are better utilized.
When applying PCA to reduce dimension of the features (\Cref{fig:pca_semseg}), we see that the \unic performance remains higher than the ones from the teachers, showing that it better utilizes the feature space.

\end{document}